\documentclass[sigconf]{acmart}
\usepackage{color}
\usepackage{algorithm}
\usepackage{algpseudocode}

\usepackage{amssymb}

\usepackage{array}
\usepackage{eqparbox}
\usepackage{stfloats}
\usepackage{url}
\usepackage{color}
\usepackage{multirow}
\usepackage{amsmath,amsfonts}
\usepackage{subfigure}
\usepackage{graphicx}
\usepackage{tikz}
\usepackage{booktabs} 

\newcommand{\pgftextcircled}[1]{
	\setbox0=\hbox{#1}%
	\dimen0\wd0%
	\divide\dimen0 by 2%
	\begin{tikzpicture}[baseline=(a.base)]%
	\useasboundingbox (-\the\dimen0,0pt) rectangle (\the\dimen0,1pt);
	\node[circle,draw,outer sep=0pt,inner sep=0.1ex] (a) {#1};
	\end{tikzpicture}
}

\newcommand{\pgftextcircledblk}[1]{
	\setbox0=\hbox{#1}%
	\dimen0\wd0%
	\divide\dimen0 by 2%
	\begin{tikzpicture}[baseline=(a.base)]%
	\useasboundingbox (-\the\dimen0,0pt) rectangle (\the\dimen0,1pt);
	\node[circle,draw,outer sep=0pt,inner sep=0.1ex,fill=blue] (a) {#1};
	\end{tikzpicture}
}

\def\t0{{t_0}}

\def\t0{{t_0}}

\newtheorem{assumption}{\textbf{Assumption}}

\newtheorem{remark}{\textbf{Remark}}
\newtheorem{examp}{\textbf{Example}}
\newtheorem{thm}{\textbf{Theorem}}
\newtheorem{prop}{\textbf{Proposition}}

\setcopyright{rightsretained}

\copyrightyear{2018}
\acmYear{2018}
\setcopyright{acmcopyright}
\acmConference[MM '18]{2018 ACM Multimedia Conference}{October 22--26, 2018}{Seoul, Republic of Korea}
\acmBooktitle{2018 ACM Multimedia Conference (MM '18), October 22--26, 2018, Seoul, Republic of Korea}
\acmPrice{15.00}
\acmDOI{10.1145/3240508.3240597}
\acmISBN{978-1-4503-5665-7/18/10}
\begin{document}
\title{A Margin-based MLE for Crowdsourced Partial Ranking}

\author{Qianqian Xu$^{1}$, Jiechao Xiong$^{2}$, Xinwei Sun$^{3,4}$, \\Zhiyong Yang$^{5}$, Xiaochun Cao$^{5}$, Qingming Huang$^{1,6,7}$$^\ast$, Yuan Yao$^{8}$$^\ast$}
\thanks{$^\ast$Corresponding author.}
\affiliation{%
 \institution{$^1$ Key Lab of Intell. Info. Process., Inst. of Comput. Tech., CAS, Beijing, 100190, China}
  \institution{$^2$ Tencent AI Lab, Shenzhen, 518057, China}
  \institution{$^3$ School of Mathematical Sciences, Peking University, Beijing, 100871, China}
  \institution{$^4$ DeepWise AI Lab, Beijing, 100085, China}
  \institution{$^5$ State Key Laboratory of Info. Security (SKLOIS), Inst. of Info. Engin., CAS, Beijing, 100093, China}
   \institution{$^6$ University of Chinese Academy of Sciences, Beijing, 100049, China}
  \institution{$^7$ Key Lab of Big Data Mining and Knowledge Management, CAS, Beijing, 100190, China}
  \institution{$^8$ Department of Mathematics, Hong Kong University of Science and Technology, Hong Kong}
}
\email{xuqianqian@ict.ac.cn,jcxiong@tencent.com,sxwxiaoxiaohehe@pku.edu.cn}
\email{{yangzhiyong,caoxiaochun}@iie.ac.cn, qmhuang@ucas.ac.cn,yuany@ust.hk}

%
%
%
%

\begin{abstract}
A preference order or ranking aggregated from pairwise comparison
data is commonly understood as a strict total order. However, in real-world scenarios, some items are intrinsically
ambiguous in comparisons, which may very well be an inherent uncertainty
of the data. In this case, the conventional total order ranking
can not capture such uncertainty with mere global ranking or utility scores. In this paper, we are
specifically interested in the recent surge in crowdsourcing applications to predict partial but
more accurate (i.e., making less incorrect statements) orders rather than complete
ones. To do so, we propose a novel framework to learn some probabilistic models of partial orders as a \emph{margin-based Maximum Likelihood Estimate} (MLE) method. We prove that the induced MLE is a joint convex
optimization problem with respect to all the parameters, including the global ranking scores and margin parameter. Moreover, three kinds of generalized linear models
are studied, including the basic uniform model, Bradley-Terry model, and Thurstone-Mosteller model, equipped with some theoretical analysis on FDR
and Power control for the proposed methods. The validity
of these models are supported by experiments with both simulated and real-world datasets, which shows that the
proposed models exhibit improvements compared with traditional
state-of-the-art algorithms.

\end{abstract}

%
%



\maketitle

\section{Introduction}
\label{Intro}

\begin{figure}[t]
	\renewcommand{\captionfont}{\footnotesize \bfseries}
	\setlength{\belowcaptionskip}{-5pt}
	\setlength{\abovecaptionskip}{0pt}
	\begin{center}
		\subfigure[Trump]{
			\includegraphics[width=0.205\textwidth]{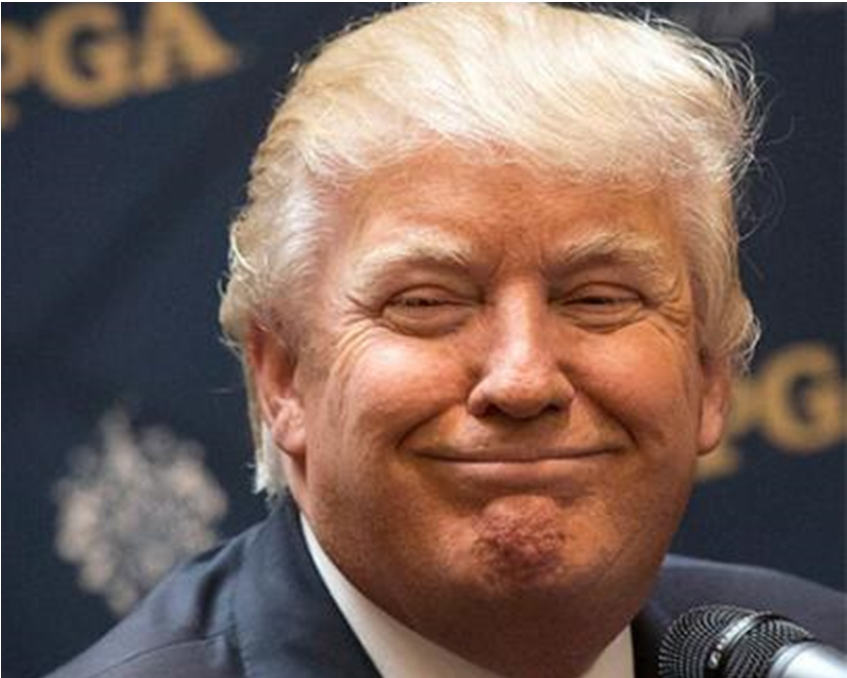}}
		\subfigure[Ming Yao]{
			\includegraphics[width=0.192\textwidth]{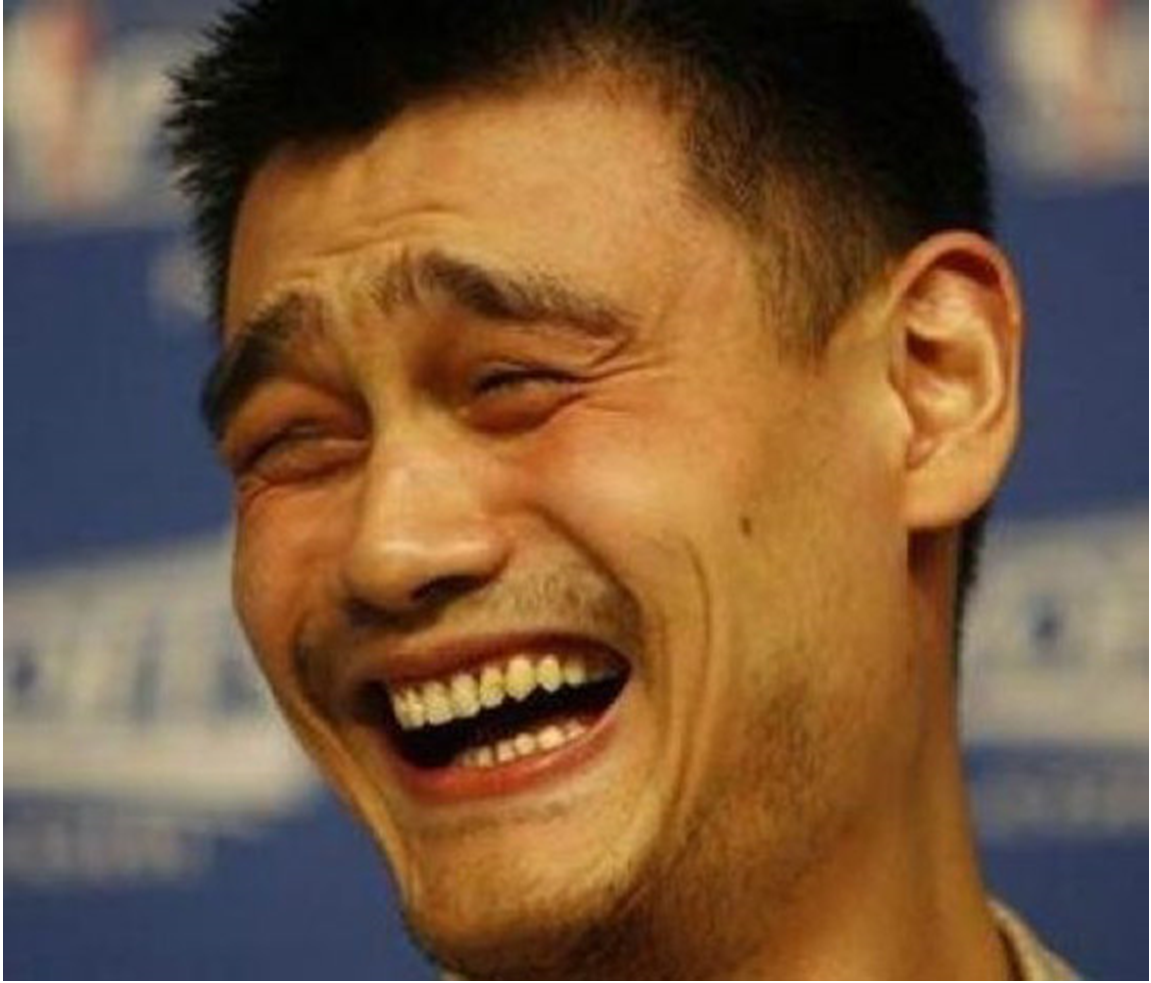}}
		\caption{Smile as a relative attribute in paired comparisons.} \label{fig:relativeattribute}
	\end{center}
\end{figure}

Imagine you are given a pile of distorted images of the same content,
and you are asked to sort or rank them according to their quality. Can you do it? In other tasks such as relative attribute ordering in computer vision, for example in Figure \ref{fig:relativeattribute}, can you rank the faces according to the ``degree" of smiling? These are typical scenarios in crowdsourced ranking.

Nature imposes a limitation that humans are unable to make accurate preference judgement
on even moderately large sets. As it has been argued that most people
can rank only between 5 to 9 alternatives at a time \cite{saaty2003magic}. This is probably why
many rating scales (e.g. the ones used by Amazon, eBay, Netflix, YouTube) are based on a 5-star (level) scale. In a 5-star test,
individuals are asked to give a rating from Bad to Excellent in 5 levels
(e.g. Bad-1, Poor-2, Fair-3, Good-4, and Excellent-5) to grade the candidates. This leads to partial orders or ranking of the candidates where the items on the same level will be regarded as equivalent classes. There are some work in the literature studying how to organize information in partial orders of such tied subsets or equivalent classes (partitions, bucket orders) \cite{gionis2006algorithms,lebanon2008non}. Specifically, the authors in \cite{lebanon2008non} address computational
aspects that arise when working with empirical distributions on partially ranked data. 

Yet in many crowdsourcing tasks, even the 5-star scale may suffer from various problems such as ambiguity in the definition
of scales, dissimilar interpretations of the scale among users, and so on, e.g. argued in \cite{MM09} and reference therein.
To address this issue, the pairwise comparison method becomes a rising paradigm recently in many crowdsourcing platforms,
as for most people, it is a harder task to rank or rate many candidates than to compare a pair of candidates at a time.
In pairwise comparisons, frequently, the available data presented to us is in
the following form: the quality of image A is better than image B, etc. A ranking aggregated from pairwise comparison data is commonly understood as a strict total order, i.e., an irreflexive, asymmetric, and transitive relation, specifying for all pairs whether $i$ precedes $j$, or $j$ precedes $i$. \cite{gionis2006algorithms} attempts to discover an underlying bucket or partial order from pairwise precedence information between the items without any ties.

Although some items or candidates could be obvious to rank, the ambiguity in choosing the preference is ubiquitous that often imposes some difficulties in making the choice. For example, the following list describes such cases met in crowdsourcing experiments.

\begin{examp}
	In relative attributes in computer vision \cite{parikh2011relative}, some attributes such as smile or age, are hard
	to judge absolutely, but accessible to human within a pair on choosing which one to be stronger in the attribute. 
Surely there might be some obvious images easy to judge. Yet there will be other images where the distinction is quite subtle, or hardly perceivable. Figure \ref{fig:relativeattribute} gives an example. Who is smiling
more, Trump or Yao? Some people may think the basketball star Ming Yao is more
smiling than Trump; while some people may think Ming Yao looks crying, so
they prefer to Trump as more smiling. Besides, others may think it is difficult to tell which one in the pair looks stronger in the
smile attribute. Participants may choose to abstain from this judgement when they are too confusing to make a decision.
\end{examp}

\begin{examp}
	In subjective multimedia quality assessment \cite{mm11,MM12}, videos and images of the same content are to be evaluated for its quality. Some pairs are easy to distinguish, while others are not. In particular, there might be multi-criteria among heterogeneous raters. In these cases, annotators may declare these two are confusing thus difficult to judge.
\end{examp}

\begin{examp}
	In crowdsourced pairwise ranking platforms such as {Allourideas}\footnote{\url{http://www.allourideas.org/}}, an option that ``I can't decide" is provided with further information such as ``I like both ideas", ``I think both ideas are the same", ``I don't like either idea", or "I don't know enough about either idea", etc. For example, in world college ranking a participant is asked about ``which university (of the following two) would you rather attend?". When a voter thinks the two colleges listed are incomparable and difficult
to judge, he may click this button with possible further options. Such voters essentially provide some information on partial orders, which can be distinguished from those voters who click this button just because they don't know both of these two colleges or one of them well.
\end{examp}

This kind of pairwise comparison data, together with ``I can't decide" type of decision, arises in a variety of crowdsourcing applications. In all these examples, if a rater is not sufficiently certain regarding the relative order of the two items, he may abstain from his choice decision and instead declare these two as being incomparable. In fact, partial ranking can be interpreted as a ranking with partial abstention. In this way, a dataset with abstention of this kind provides us information about possible ties or equivalent classes of items in partial orders.


Despite a considerable amount of work on ranking in general and pairwise ranking in particular, there lacks a systematic treatment on learning partial orders or rankings from such pairwise comparison data with abstentions, which are ubiquitous in crowdsourcing applications nowadays. Among the prior work on partial ranking up to our knowledge, the one that comes closest to our goal is \cite{cheng2010predicting}. The idea is that it produces predictions in the form of partial order by thresholding a (valued) pairwise preference relation, i.e., by a ``$\alpha$-cut" of preference relation. However, it leaves the optimal choice of hyper-parameter $\alpha$ to various heuristics and needs to know in advance the preference relation between every pair of items (i.e., $n(n - 1)/2$ pairs in total for $n$ items), which requires a large number of comparisons, being too prohibitive in modern applications.

To fill in this gap, in this paper, we propose a novel framework to learn partial ranking probabilistic models as a margin-based Maximum Likelihood Estimate (MLE) method. In contrast to \cite{cheng2010predicting}, all the parameters, including the global ranking score and the hyper parameter as threshold (called margin parameter here), can be automatically learned from pairwise comparison data with abstentions via a convex optimization. Our framework can deal with incomplete and imbalanced data, as an extension of the HodgeRank \cite{hodge} from total orders to partial orders with generalized linear models.

%
%
%


As a summary, our main contributions in this new framework are highlighted as follows:
\begin{itemize}
	\item[(A)] We propose a framework of learning partial rankings from pairwise comparison data with abstentions, based on a margin-based Maximum Likelihood Estimate (MLE) for probabilistic models.  We prove that for a general class of models, the induced MLE is a convex optimization problem with respect to all the parameters, including the global ranking scores and threshold/margin parameter.
	\item[(B)] In this unified framework, three kinds of generalized linear models are particularly studied, including the basic uniform model, Bradley-Terry model, and Thurstone-Mosteller model, equipped with theoretical analysis on FDR and Power control of our proposed method.
	\item[(C)] Experiments on simulated and crowdsourcing real-world datasets together show that our algorithm works effectively in practice.
\end{itemize}

The remainder of this paper is organized as follows. Sec.\ref{sec:relatedwork} contains a review of related work. We systematically introduce the methodology for partial ranking in Sec.\ref{sec:methodology}. Detailed experiments with simulated and real-world datasets are presented in Sec.\ref{sec:experiments}. Finally, Sec.\ref{sec:conclusions} presents the conclusive remarks.

\section{Related work} \label{sec:relatedwork}

\textbf{Pairwise Ranking}. Statistical preference aggregation, in particular ranking or rating from pairwise comparisons, is a classical problem that can be traced back to the $18^{th}$ century. This subject area has been widely studied in various fields including the social choice and voting theory in economics~\cite{Condorcet,Arrow51}, statistics~\cite{Noether60, David88},  multimedia \cite{mm11,MM12}, computer vision~\cite{Yu09,Yu12,Osher11_retinex}, and others~\cite{Pagerank,Thurstone27,Saaty77,Hits,Stefani77,Elo++,osting2013statistical,ICML14,CorMohRas07}.

Various algorithms have been studied to solve this problem. They include maximum likelihood under a Bradley-Terry model assumption, rank centrality (PageRank/MC3) \cite{negahban2012,dwork2001rank}, HodgeRank~\cite{hodge}, and a pairwise variant of Borda count \cite{de1781memoire}, etc.
However, all of these methods have a major drawback: they aim to find one global common consensus, that assumes
all users' choices are stochastic revelation of a common global
preference function on candidates. To capture the discrepancies among users, lately \cite{xu2016parsimonious} proposes a parsimonious mixed effect HodgeRank, which considers that a majority of users may follow
the common social preferences while some users may exhibit distinct personalized preferences. However, all these methods above do not
consider the inherent characteristic of real-world data: some pairs are intrinsically ambiguous, thus may be difficult to derive a strict global ranking. In this paper, we will focus on this kind of setting, allowing a model to make predictions in the form of
partial instead of total orders.

\textbf{Partial Ranking}. Despite a considerable amount of work on ranking in general and pairwise ranking, the
literature on partial rankings is relatively sparse. Pairwise comparisons with abstentions are governed by partial orders or rankings. But the notion of abstention is actually originated from classification community \cite{Chow1970}. In classification with a reject option, for example, a classifier
may abstain from a class prediction if making no decision is considered less harmful than making an
unreliable and hence potentially false decision. Recently, worth mentioning is the work on a specific type of
partial orders, namely linear orders of unsorted or tied subsets (partitions, bucket orders) \cite{gionis2006algorithms,lebanon2008non}.
However, the problems addressed in these
work are different from our goals. Among the existing work in the literature,
\cite{cheng2010predicting} is the one that comes closest to our goal, which produces predictions in the form of partial order by thresholding a (valued) pairwise
preference relation, i.e., by a ``$\alpha$-cut" of preference relation. It lacks a solid principle to decide the hyper parameter $\alpha$ as the threshold. Moreover, it needs to know in advance the preference relation between every pair of items.
In this paper, we propose a margin-based MLE for partial order ranking based on probability model which could solve these problems in \cite{cheng2010predicting}.

\section{Methodology}\label{sec:methodology}
\subsection{Pairwise Ranking on Graphs}

Suppose there are $n$ alternatives or items to be ranked. The pairwise comparison
labels collected from users
can be naturally represented as a directed comparison
graph $G = (V;E)$. Let $V = \{1,2,\dots,n\}$ be the vertex set of $n$ items and $E = \{(u,i,j): i,j\in V, u \in U\}$ be the set of edges, where $U$ is the
set of all users who compared items. User $u$ provides his/her preference between choice $i$ and $j$,  such that $y_{ij}^u>0$ means $u$
prefers $i$ to $j$ and $y_{ij}^{u}\leq 0$ otherwise. Hence we may assume $y: E\rightarrow R$ with skew-symmetry
(orientation) $y_{ij}^u=-y_{ji}^u$. The magnitude of $y_{ij}^u$ can represent the degree of preference and it varies
in applications. The simplest setting is the binary choice, where $y_{ij}^u = 1$ if $u$ prefers $i$ to $j$ and $y_{ij}^u=-1$ otherwise.

Traditionally, a statistical ranking is commonly understood as a strict total order, i.e., an irreflexive,
asymmetric, and transitive relation $>$, specifying for all pairs whether $i$ precedes $j$, denoted $i > j$, or $j$ precedes $i$. The key property of transitivity can be seen as a principle of consistency: If $i$ is preferred to $j$ and $j$ is preferred to $k$, then $i$ must be preferred to $k$. However, in real-world applications, some pairs are intrinsically
ambiguous, in this case, the rater cannot reliably
decide whether the former should precede the latter or the other way around,
he may abstain from this decision and instead declare these alternatives as
being incomparable. Therefore,
it might be misleading to merely look at a global total ranking (i.e., in which every pair of distinct
elements is comparable) while ignoring the intrinsic ambiguity among items.
In this paper, we focus on deriving a partial ranking based on a margin-based MLE method.

\subsection{Partial Order Ranking}
A partial order $\succ$ is a generalization of the relation $>$ mentioned above that preserves the consistency principle but is not necessarily total. Define $i \succ j$ as $ (i>j)\wedge(^\urcorner (j > i))$ . If, for two alternatives $i$ and $j$, neither
$i \succ j$ nor $j \succ i$, then these alternatives are considered as incomparable, we then denote $i ~\bot~ j$ or equivalently $j ~\bot~ i$.
In other words, if $i$ and $j$ are too similar such that we think neither $i$ precedes $j$ nor $j$ precedes $i$, we then claim that $i$ and $j$ are incomparable.

In \cite{cheng2010predicting}, it proposes to learn a Partial Order Relation (POR) by a ``$\alpha$-cut" of preference relation. Suppose $P(i,j)$ is a measure of support for the order (preference) relation $i\succ j$ with property $P(j,i) = 1 - P(i,j)$. Then a POR is defined as
\[\mathcal{R}_\alpha = \{i\succ j:P(i,j)\ge\alpha\}\]
by setting $\alpha$ big enough.

However, this construction of POR requires the preference relation between every pair of items (i.e., $n(n-1)/2$ pairs in total for $n$ items). And each $P(i,j)$ is usually estimated by empirical probability between $i$ and $j$. Therefore, a good estimation of $P(i,j)$ needs a large number of comparisons.

\subsection{Probability Model for Binary Data}
In order to extend the methods to the case of small number of samples, we introduce the probability model for binary data. Suppose that the true scaling scores for $n$ items are $\boldsymbol{s} = [s_1,\cdots, s_n]$ and we collect $N$ pairwise comparison samples $\{(i_k,j_k,y_{k})\}_{k=1}^N$ in total. Here $(i_k, j_k)$ is a pair of items, and $y_k$ is the corresponding comparison label. Suppose that, for the $k$th
observation, $y_k$ is generated by:
\[y_{k} = \mathrm{sign}(s_{i_k}-s_{j_k}+\epsilon_{k}),\]
where $\epsilon_{k}$ are $i.i.d$ and have a c.d.f $\Phi(t)$. Different $\Phi$ leads to different models. For example:
\begin{itemize}
	\item Uniform model: $\Phi(t) = \frac{t+1}{2}$.
	\item Bradley-Terry model: $\Phi(t) = \frac{e^t}{1+e^t}$.
	\item Thurstone-Mosteller model: $\Phi(t) = \frac{1}{\sqrt{2\pi}}\int_{-\infty}^{x}e^{-\frac{t^2}{2}}dt$.
	
\end{itemize}

Note that $P(y_{k} = 1) = 1-\Phi(s_{j_k}-s_{i_k})$, a $\alpha$-cut of preference relation is thus equivalent to a $-\Phi^{-1}(1-\alpha)$-cut of the score difference function $f(k) = s_{i_k}-s_{j_k}$. Therefore, a POR can be obtained if the score can be estimated, which allows the comparison samples to be incomplete. It can be proved, such a cut indeed implies a POR.
\begin{prop}
For any $\boldsymbol{s}$ and $\lambda>0$, the relation
\begin{equation}\label{POR}
 \mathcal{R}_\lambda= \{i\succ j:s_i-s_j>\lambda\}
\end{equation}
is a Partial Order.
\end{prop}

\subsection{Extended Probability Model }
As stated that, any value of $\lambda$ can produce a POR, then how can we choose a proper one is a key step. In real-world applications, as some pairs are intrinsically ambiguous, raters usually provide a third option, ``I can't decide" or ``They are comparable". Such kind of data can help us to determine the ``optimal cut" here.
To fit this kind data with three options, we extend the probability model as follows:
\begin{equation}\label{ProbModel}
y_{k} = \left\{
\begin{array}{l l }
1,&s_{i_k}-s_{j_k}+\epsilon_{k} >\lambda;\\
-1,&s_{i_k}-s_{j_k}+\epsilon_{k} <-\lambda;\\
0,&\mbox{else} .
\end{array}\right.
\end{equation}
where $y_{k} = 0$ indicates the annotator thinks that $i$ and $j$ are too close to judge. Then under this model, the POR in \eqref{POR} has an explicit meaning: an oracle annotator, whose $\epsilon_{ij} = 0$, will give exactly this POR!

\subsection{Maximum Likelihood Estimator}\label{sec:MLE}
With the label distribution modeled, in this section we elaborate a Maximum Likelihood method to estimate the model parameters. First, we construct the design matrix $\boldsymbol{X}$ as $\boldsymbol{X} = [\boldsymbol{x}_1^\top, \cdots \boldsymbol{x}_N^\top]^\top$, where $\boldsymbol{x}_k = \boldsymbol{e}_{j_k} - \boldsymbol{e}_{i_k} $. Furthermore, we denote $\theta = [\lambda, \boldsymbol{s}]$ . With the notations above, we could calculate the possibility that $y_k =1,0,-1$ as follows:
\begin{align*}
P\{y_k=1\} &= P\{\epsilon_{k} > \lambda - s_{i_k} +s_{j_k} \} =1-  \Phi\big([1, \boldsymbol{x}_k^\top]^\top \boldsymbol{\theta} \big), \nonumber \\
P\big\{y_k=0\big\} &= P\{ -\lambda - s_{i_k} +s_{j_k}<\epsilon_{k} \le \lambda - s_{i_k} +s_{j_k} \}, \nonumber \\ &= \Phi\big([1, \boldsymbol{x}_k^\top]^\top \boldsymbol{\theta}\big) - \Phi\big([-1, \boldsymbol{x}_k^\top]^\top\boldsymbol{\theta} \big), \nonumber \\
P\{y_k=-1\} &= P\{\epsilon_{k} \le -\lambda - s_{i_k} +s_{j_k} \} = \Phi\big([-1, \boldsymbol{x}_k^\top]^\top \boldsymbol{\theta} \big). \nonumber
\end{align*}
Therefore:
\[
P\{y_k\} =\prod_{label \in\{-1,0,1\}}
\Bigg[P\{y_k=label\}\Bigg]^{1\{y_k=label\}}.
\]
Given all above, it is easy to write out the negative  log-likelihood via
denoting $\zeta_k^+$ as $[1, \boldsymbol{x}_k^\top]^\top\boldsymbol{\theta} $
and
$\zeta_k^-$ as $[-1, \boldsymbol{x}_k^\top]^\top\boldsymbol{\theta} $
\begin{equation}\label{likelihood}
\begin{split}
&\ell(y|\boldsymbol{s},\lambda)  = -\sum_{k}\bigg(1\{y_{k}=1\} log\Big[1-\Phi(\zeta_k^+)\Big]\\ &+ 1\{y_k=0\}log\Big[\Phi(\zeta_k^+) - \Phi(\zeta_k^-)\Big]
+ 1\{y_k=-1\} log\Big[\Phi(\zeta_k^-)\Big]\bigg).
\end{split}
\end{equation}
To solve our proposed model, one just needs to minimize $\ell(y|\boldsymbol{s}, \lambda)$ with respect to $(\lambda, \boldsymbol{s})$.
Furthermore, if we assume that $\sum_{i}s_i =0$, then we could replace $s_n$ with $-\sum_{i=1}^{n-1} s_{i}$. Correspondingly, we can rewrite the loss function $\ell(y | \boldsymbol{s},\lambda)$ as a function of $(\lambda,s_{1},...,s_{n-1})$: ${\ell}(y | \boldsymbol{s}/s_{n}, \lambda)$ , where $\boldsymbol{s}/s_{n} \triangleq (s_{1},...,s_{n-1})$.
Then, under Assumption \ref{asumm} we could prove that Theorem 3.2 holds.
\begin{assumption} \label{asumm}
   Define $\phi(x)$ as $\Phi'(x)$, we assume that at least one of the
   following assumptions holds for $\Phi(x), \phi(x)$ and $\phi'(x)$:
	\begin{itemize}
		\item[a)] $\phi'(x)  \equiv  0 $ and $\forall x, \phi(x) \neq 0 $;
		\item[b)] $\phi(x)$ is an even  function, $\phi(x)$ and $\Phi(x)-
		\frac{\phi^2(x)}{\phi(x)}$ is strictly decreasing on $(0, +\infty)$.
		Furthermore, $\lim\limits_{x \rightarrow +\infty}
		\frac{\phi^2(x)}{\phi'(x)}=0$, $\phi'(x) \neq 0$ for $x \neq 0$,
		$\forall x, \phi(x) \neq  0$ and $\phi'(x) < 0$ if $x > 0$ .	
	\end{itemize}

\end{assumption}
\begin{thm}\label{cvx}
	${\ell}(y | \boldsymbol{s}/s_{n}, \lambda)$ is strictly convex with respect to $(\lambda, s_1$, $\cdots, s_{n-1})$.
\end{thm}
It is easy to check all the three models satisfy Assumption \ref{asumm}, thus all these models are strictly convex.

Putting all these together, we conclude that the MLE of these models are just solutions of strictly convex problems which can be solved efficiently.
\subsection{FDR and Power Control}
In this part, we show the theoretical analysis of model performance on separating the incomparable pairs from the comparable
ones. To measure this ability, we introduce two criteria: False Discovery Rate (FDR) and Power, as is shown in Table \ref{tab:power}.
\begin{table}
 \renewcommand{\captionfont}{\footnotesize \bfseries}
\caption{The definition of FDR and Power.}
\centering
	\small
	\begin{tabular}{|c|c|c|c|c|c|c|c|c|c|c|}
		\hline
		&  Comparable   & Incomparable   \\
		\hline
		Detected as Comparable  & $\mathcal{N}_{0,0}$ & $\mathcal{N}_{0,1}$ \\
		\hline
		Detected as Incomparable  & $\mathcal{N}_{1,0}$ & $\mathcal{N}_{1,1}$ \\
		\hline
	\end{tabular}\label{tab:power}
\end{table}
The definition of FDR and Power in our setting are:
\begin{align}
FDR & = \frac{\mathcal{N}_{1,0}}{\mathcal{N}_{1,0} + \mathcal{N}_{1,1}},
\nonumber \\
Power & = \frac{\mathcal{N}_{1,1}}{\mathcal{N}_{0,1} + \mathcal{N}_{1,1}}.
\nonumber
\end{align}
In the following, we will propose a conservative threshold bound to guarantee FDR to be 0 and an
aggressive threshold bound to guarantee Power to be 1. \\
\indent  Let $\big(\{s^\star_i\}_{i=1}^n, \lambda^\star \big)$ be the
corresponding true parameters, $\big(\{\hat{s}_i\}_{i=1}^n$, $\hat{\lambda}\big)$
be the corresponding estimated parameters returned by our proposed method.
Denote $\delta$ as the maximum of the variance of the estimated model
parameters i.e. $\delta = max({\sigma^2_{\hat{\lambda}},\sigma^2_{\hat{s}_1},
\cdots,
\sigma^2_{\hat{s}_n}})$. Furthermore, we denote $\hat{\delta}$ as the
	estimation of $\delta$
on the observed dataset and $\Delta =\frac{\sqrt{4log(n+1)\hat{\delta}}}{\sqrt{N}} $. With the notations above, we
construct the set of all incomparable pairs as $\mathcal{M}$, a
conservative set as ${\mathcal{\widehat{M}}}$ and the aggressive set as
$\mathcal{\widetilde{M}}$ :
\begin{align}
\mathcal{M} & =  \{(i,j) : \vert {s}^\star_{i} - {s}^\star_{j} \vert \leq
\lambda^* \}, \\
{\mathcal{\widehat{M}}} &= \{(i,j) : \vert \hat{s}_{i} - \hat{s}_{j} \vert \leq
\hat{\lambda} - 3\Delta \}, \\
\mathcal{\widetilde{M}} &= \{(i,j) : \vert \hat{s}_{i} - \hat{s}_{j} \vert \leq
\hat{\lambda} + 3\Delta\},
\end{align}
where  $N$ is the number of samples. Now we first propose a theorem which shows that with high probability,  ${\mathcal{\widehat{M}}} \subseteq \mathcal{M} \subseteq\mathcal{\widetilde{M}}$, followed by a practical interpretation via the remark that comes right after the theorem.\newline

\begin{thm}\label{bound}
	Let $\theta = (\lambda,\boldsymbol{s})$. Then with probability at least $1 - 2(n+1)^{\frac{\delta - 2\hat{\delta}}{\delta} }$, we will have that ${\mathcal{\widehat{M}}} \subseteq \mathcal{M} \subseteq \mathcal{\widetilde{M}}$.
\end{thm}
\begin{remark}
	If  $\mathcal{\widehat{M}} \subseteq \mathcal{M}$ occurs, we set the threshold
	$\lambda$ as $\underline{\lambda} = \hat{\lambda}-3\Delta$. Then all the
	detected incomparable pairs are truly incomparable, thus FDR = 0 is guaranteed. Likewise, if $\mathcal{{M}} \subseteq
	\mathcal{\widetilde{M}}$ i.e. $\mathcal{\widetilde{M}}^c \subseteq
	\mathcal{{M}}^c$ occurs, we have $\vert \hat{s}_i - \hat{s}_j\vert >\hat{\lambda} + 3\Delta $ indicating $\vert {s}^\star_i - {s}^\star_j\vert >\lambda^\star $. Consequently, if we set the threshold as $\overline{\lambda} = \hat{\lambda} +
	3\Delta$, then all the comparable pairs will be detected as comparable and thus
	Power = 1 is guaranteed.
\end{remark}
To evaluate $\overline{\lambda}$ and $\underline{\lambda}$, one must first evaluate
$\hat{\delta}$. Next, we propose a method to estimate $\hat{\delta}$ with the
well-known asymptotic normality of MLE \cite{asymp}. First, according to
Section \ref{sec:MLE}, we know that
${\ell}(y |\boldsymbol{s}/s_{n},
\lambda)$ is strictly convex for all mentioned distributions. Denote $\tilde{I}((\hat{\lambda},\hat{\boldsymbol{s}}/\hat{s}_{n}))$ as the estimated
Fisher Information matrix,  we have:
\begin{align}
\tilde{I}((\hat{\lambda},\hat{\boldsymbol{s}}/\hat{s}_{n})) =
-\triangledown^{2}_{\lambda,\boldsymbol{s}/s_{n}}\Big[{\ell}(y |
(\hat{\lambda},\hat{\boldsymbol{s}}/\hat{s}_{n}))/N\Big] \succ 0 \nonumber,
\end{align}
and
\begin{align}
E\Big[\frac{-\nabla^{2}_{\lambda,\boldsymbol{s}/s_{n}}{\ell}(y |
\boldsymbol{s}^{\star}/s^{\star}_{n},\lambda^{
	\star})}{N}\Big] =
	I((\lambda^{\star},\boldsymbol{s}^{\star}/s^{\star}_{n})) \succ 0, \nonumber
\end{align}
where $I((\lambda^{\star},\boldsymbol{s}^{\star}/s^{\star}_{n}))$ is the true
Fisher Information matrix. Hence, these two matrices are invertible while the
inversion has positive diagonal elements.
Accordingly we have:
\begin{align}
I^{-1}((\lambda^{\star},\boldsymbol{s}^{\star}/s^{\star}_{n}))_{1,1} & =
\sigma^{2}_{\hat{\lambda}}; \nonumber \\
I^{-1}((\lambda^{\star},\boldsymbol{s}^{\star}/s^{\star}_{n}))_{i,i} & =
\sigma^{2}_{\hat{s_{i}}}, \quad \forall i=1,...,n-1. \nonumber
\end{align}
Then, we could estimate the variances as:
\begin{align}
\hat{\sigma}^{2}_{\hat{\lambda}} & \triangleq
\tilde{I}^{-1}((\hat{\lambda},\hat{\boldsymbol{s}}/\hat{s}_{n}))_{1,1};
\nonumber \\
\hat{\sigma}^{2}_{\hat{s}_{i}} & \triangleq
\tilde{I}^{-1}((\hat{\lambda},\hat{\boldsymbol{s}}/\hat{s}_{n}))_{i,i}, \quad
\forall i=1,...,n-1; \nonumber \\
\hat{\sigma}^{2}_{\hat{s}_{n}} & \triangleq
(0,1,1,..,1)\tilde{I}^{-1}((\hat{\lambda},\hat{\boldsymbol{s}}/\hat{s}_{n}))(0,1,1,...,1)^{\top}.
  \nonumber
\end{align}
Similarly, we can estimate $\hat{\delta}$ as:
\begin{align}
\hat{\delta} = max\{\hat{\sigma}^{2}_{\hat{\lambda}},\hat{\sigma}^{2}_{\hat{s}_{1}},...,\hat{\sigma}^{2}_{\hat{s}_{n}}\}. \nonumber
\end{align}

\section{Experiments}\label{sec:experiments}

In this section, four examples are exhibited with both
simulated and real-world data to illustrate the validity of
the analysis above and applications of the methodology proposed.
The first example is with simulated data while the
latter three exploit real-world data collected by crowdsourcing.
\subsection{Simulated Study}
\textbf{Settings} We validate the proposed algorithm on simulated data with $n=|V|=20$ labeled by users.
Specifically, we first randomly create a global ranking score $\boldsymbol{s}^{\star} \sim 10 \times N(0,1)$ as the ground-truth for $n$ candidates. Then pairwise comparisons are generated by Bradley-Terry model, i.e. $y_{i,j} = 1$ with probability $\Big\{\frac{exp(s^{\star}_{i} - s^{\star}_{j} - \lambda^{\star})}{1 + exp(s^{\star}_{i} - s^{\star}_{j} - \lambda^{\star})} \Big\}$, $y_{i,j} = 0$ with probability $\Big\{\frac{exp(s^{\star}_{i} - s^{\star}_{j} + \lambda^{\star})}{1 + exp(s^{\star}_{i} - s^{\star}_{j} + \lambda^{\star})} - \frac{exp(s^{\star}_{i} - s^{\star}_{j} - \lambda^{\star})}{1 + exp(s^{\star}_{i} - s^{\star}_{j} - \lambda^{\star})} \Big\}$, and $y_{i,j} = -1$ with probability $\Big\{\frac{1}{1 + exp(s^{\star}_{i} - s^{\star}_{j} + \lambda^{\star})} \Big\}$. Here we set $\lambda = 0.5:0.5:2$. Finally, we obtain a dataset with 10000 samples. The experiments are repeated 20 times and ensemble statistics for the estimator are recorded. \\
\textbf{Evaluation metrics} We measure the experimental results via two
evaluation criteria, i.e., Macro-F1, and
Micro-F1, which take both precision and recall into
account. The larger the value of Micro-F1 and Macro-F1,
the better the performance. More details about the
evaluation metric please refer to \cite{zhou2014}.\\
\textbf{Results} Table \ref{tab:macrof} and \ref{tab:microf} show the Macro-F1 and Micro-F1 of three models with $\lambda = 1$. Since the observed dataset is generated from the Bradley-Terry model, it obtains the best performance in terms of both metrics. Moreover, we also show the experimental results as $\lambda$ varies in Table \ref{tab2:macrof} and \ref{tab2:micro}, and it is easy to find that Bradley-Terry model again exhibits the best performance in most cases. In the following real-world datasets, we will also show the experimental results of Bradley-Terry Model.\\
\textbf{Validation of the FDR and Power guarantee.} To demonstrate the correctness of Theorem \ref{bound}, we plot the FDR and Power results in Figure \ref{fig:bound} for $\lambda = 0.25:0.25:2$ when $\hat{\lambda}$, $\hat{\lambda} -3 \Delta$ and $\hat{\lambda} +3 \Delta$ are employed as the estimated threshold, respectively. From the results we can easily find that, when $\hat{\lambda} - 3\Delta$ is employed as the estimated threshold, the FDR could always reach 0; while $\hat{\lambda} + 3\Delta$ could ensure the Power to be 1. This observation effectively demonstrates the correctness of the constructed conservative/aggressive set for FDR/Power. \\
{\renewcommand\baselinestretch{1.0}\selectfont

	\begin{table} [t]
 \renewcommand{\captionfont}{\footnotesize \bfseries}
  \setlength{\belowcaptionskip}{-10pt}
	 \setlength{\abovecaptionskip}{0pt}
	\caption{\label{runtime} Experimental results of 3 models on simulated data ($\lambda = 1$).}
	
	\centering
	
	\subtable[Macro-F1]{
		\newsavebox{\tablebox}
		\begin{lrbox}{\tablebox}
			\begin{tabular}{lllll}
				\hline     &min  &mean &max &std\\
				\hline
				\hline  \textbf{Uniform}       &0.7842    &\textbf{\textbf{0.8454}}    &0.9632    &0.0437 \\
				\hline  \textbf{Bradley-Terry}     &0.8309    &\textbf{\textbf{0.9794}}    &1.0000    &0.0265 \\
				\hline  \textbf{Thurstone-Mosteller}      &0.8747    &\textbf{\textbf{0.9679}}    &1.0000    &0.0312 \\
				\hline
				\end {tabular}
			\end{lrbox}
			\scalebox{0.85}{\usebox{\tablebox}}
			\label{tab:macrof}
		}

		\subtable[Micro-F1]{
			\begin{lrbox}{\tablebox}
				\begin{tabular}{lllll}
					\hline     &min  &mean &max &std\\
					\hline
					\hline  \textbf{Uniform}      &0.7872    &\textbf{\textbf{0.8611}}    &0.9677    &0.0389 \\
					\hline  \textbf{Bradley-Terry}     &0.8214    &\textbf{\textbf{0.9803}}    &1.0000    &0.0263 \\
					\hline  \textbf{Thurstone-Mosteller}    &0.8908    &\textbf{\textbf{0.9749}}    &1.0000    &0.0260 \\
					\hline
					\end {tabular}
				\end{lrbox}
				\scalebox{0.85}{\usebox{\tablebox}}
				\label{tab:microf}
			}

	\end{table}
	\par}

{\renewcommand\baselinestretch{1.0}\selectfont
	
	\begin{table} [t]
 \renewcommand{\captionfont}{\footnotesize \bfseries}
 \setlength{\belowcaptionskip}{-10pt}
	 \setlength{\abovecaptionskip}{0pt}
		\caption{ Experimental results of 3 models on simulated data as $\lambda$ varies ($\lambda = 0.5, 1, 1.5, 2$).}
		
		\centering
		
		\subtable[Macro-F1]{
			\begin{lrbox}{\tablebox}
				\begin{tabular}{c||p{0.7cm}p{0.7cm}p{0.7cm}p{0.7cm}}
					\hline   $\lambda$  &0.5  &1 &1.5 &2\\
					\hline
					\hline  \textbf{Uniform}     &0.8017    &0.8454    &0.8369    &0.8068 \\
					\hline  \textbf{Bradley-Terry}     &0.9753    &0.9794    &0.9761    &0.9818 \\
					\hline  \textbf{Thurstone-Mosteller}    &0.9628    &0.9679    &0.9714    &0.9727 \\
					\hline
					
					\end {tabular}
				\end{lrbox}
				\scalebox{0.85}{\usebox{\tablebox}}
				\label{tab2:macrof}
			}

		\subtable[Macro-F1]{
			\begin{lrbox}{\tablebox}
				\begin{tabular}{c||p{0.7cm}p{0.7cm}p{0.7cm}p{0.7cm}}
					\hline   $\lambda$  &0.5  &1 &1.5 &2\\
					\hline
					\hline  \textbf{Uniform}     &0.8520    &0.8611    &0.8273    &0.7987 \\
					\hline  \textbf{Bradley-Terry}      &0.9794    &0.9803    &0.9761    &0.9814\\
					\hline  \textbf{Thurstone-Mosteller}     &0.9822    &0.9749    &0.9710    &0.9704 \\
					\hline
					
					\end {tabular}
				\end{lrbox}
				\scalebox{0.85}{\usebox{\tablebox}}
				\label{tab2:micro}
			}
	
\end{table}
\par}

\begin{figure}[t]
	\renewcommand{\captionfont}{\footnotesize \bfseries}
	 \setlength{\belowcaptionskip}{-10pt}
	\setlength{\abovecaptionskip}{5pt}
	\begin{center}
		\subfigure[FDR]{\includegraphics[width=0.22\textwidth]{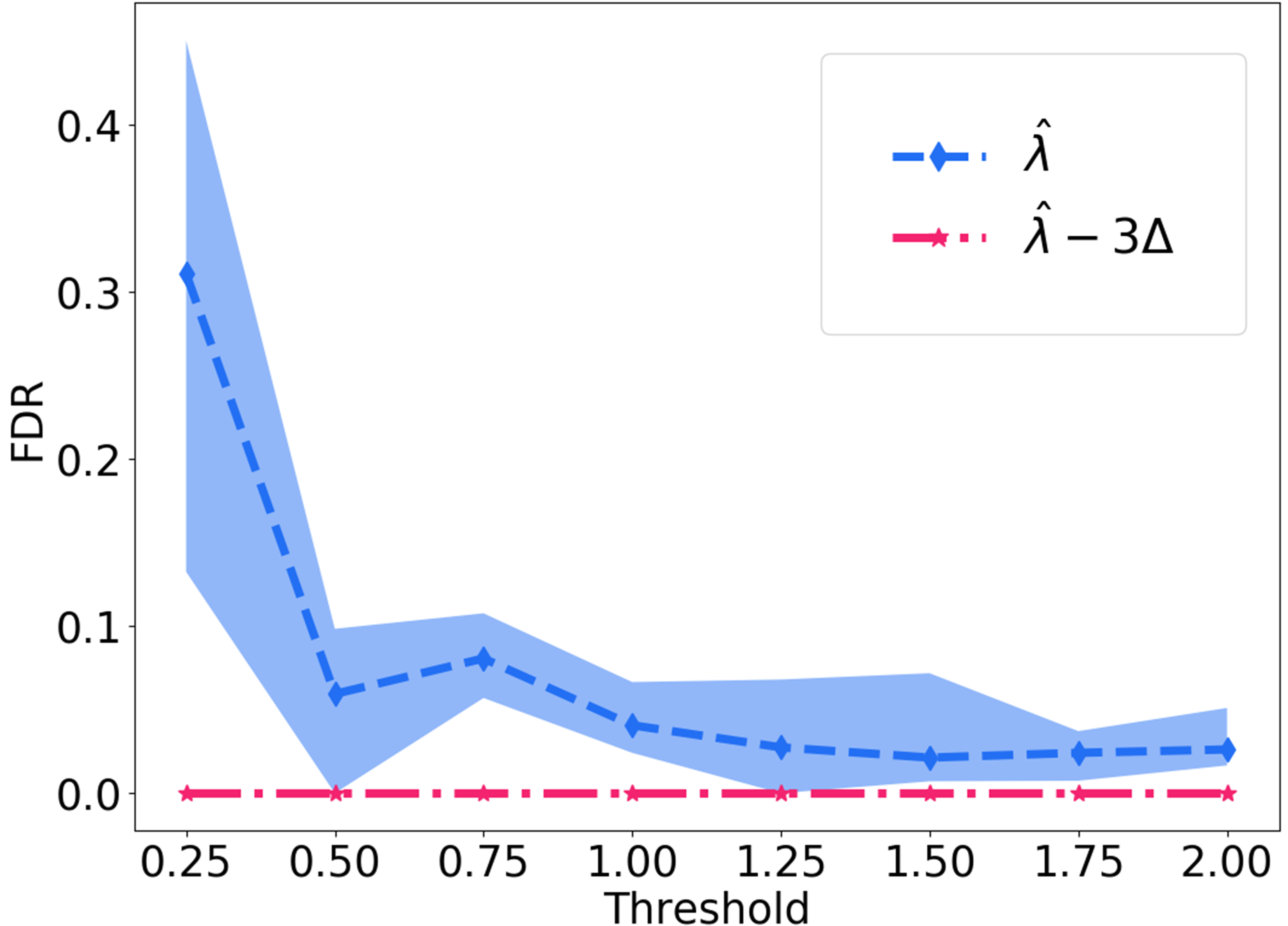}}
		\subfigure[Power]{
			\includegraphics[width=0.22\textwidth]{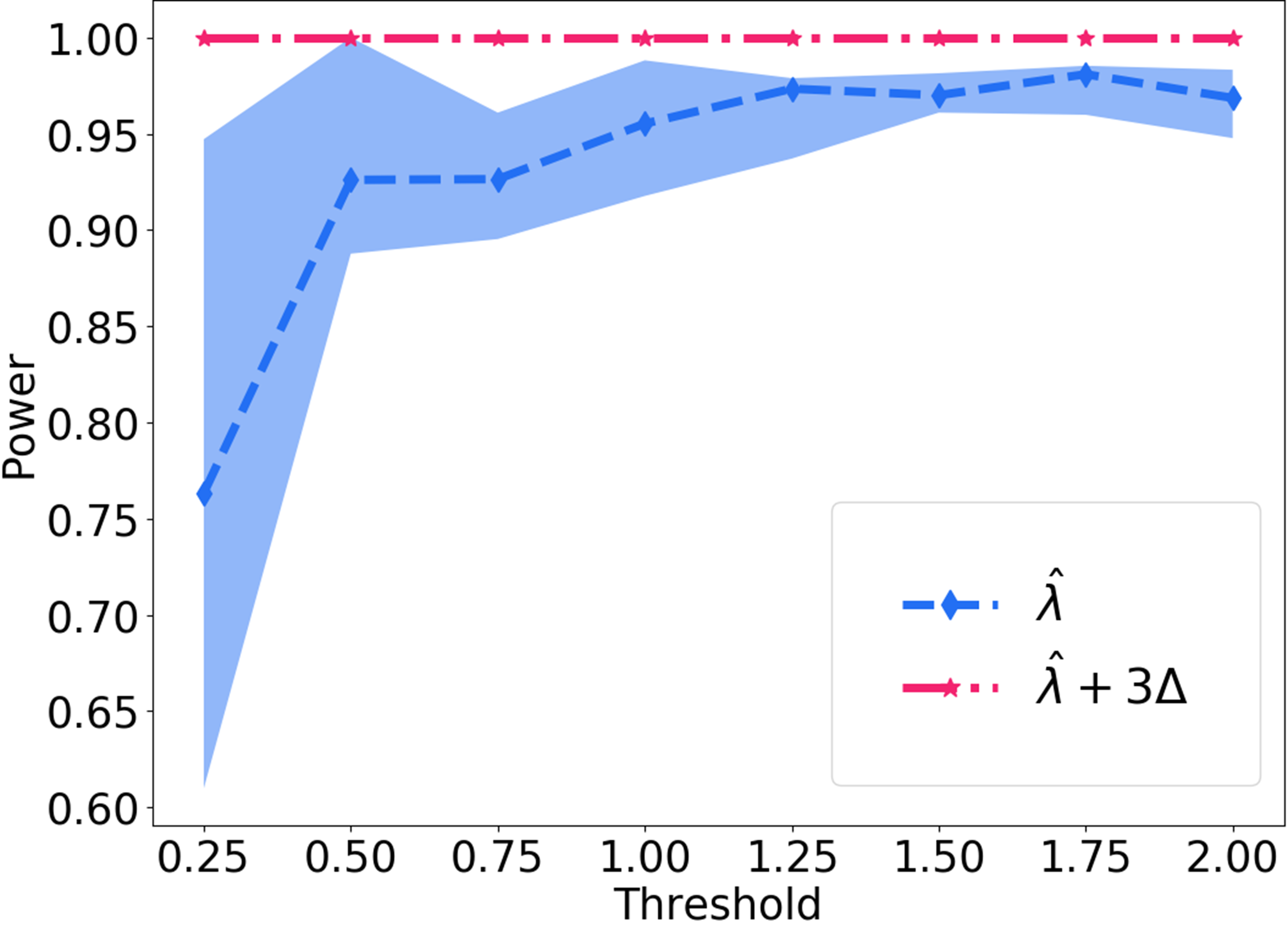}}
		\caption{An illustration of FDR and Power control.} \label{fig:bound}
	\end{center}
\end{figure}

%
%
%
%

\begin{figure}[t]
	 \renewcommand{\captionfont}{\footnotesize \bfseries}
	\setlength{\belowcaptionskip}{-5pt}
	\begin{center}
		\subfigure[Optimal $\lambda$]{
			\includegraphics[width=0.27\textwidth]{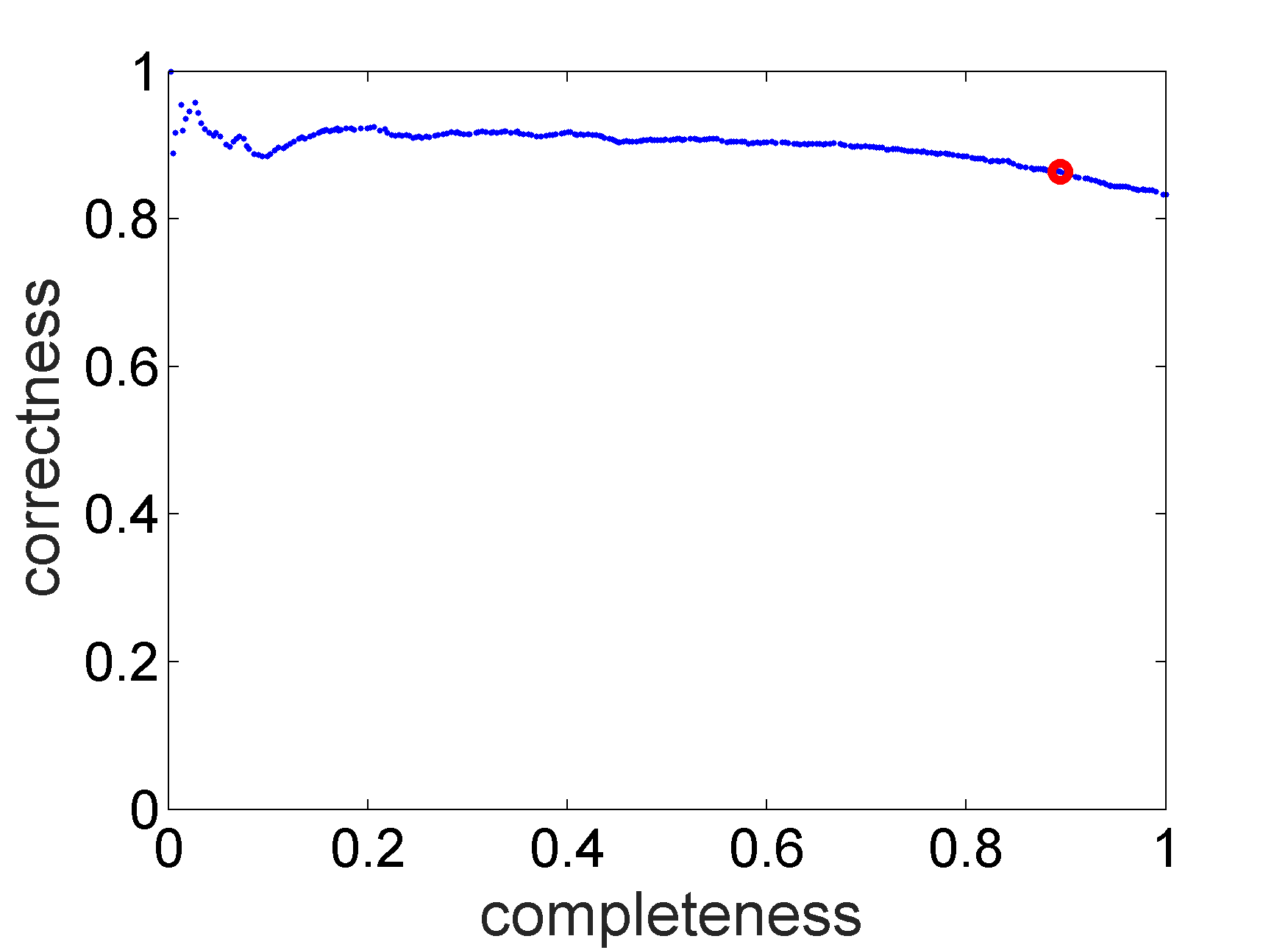}}
		\subfigure[Partial ranking]{
			\includegraphics[width=0.13\textwidth]{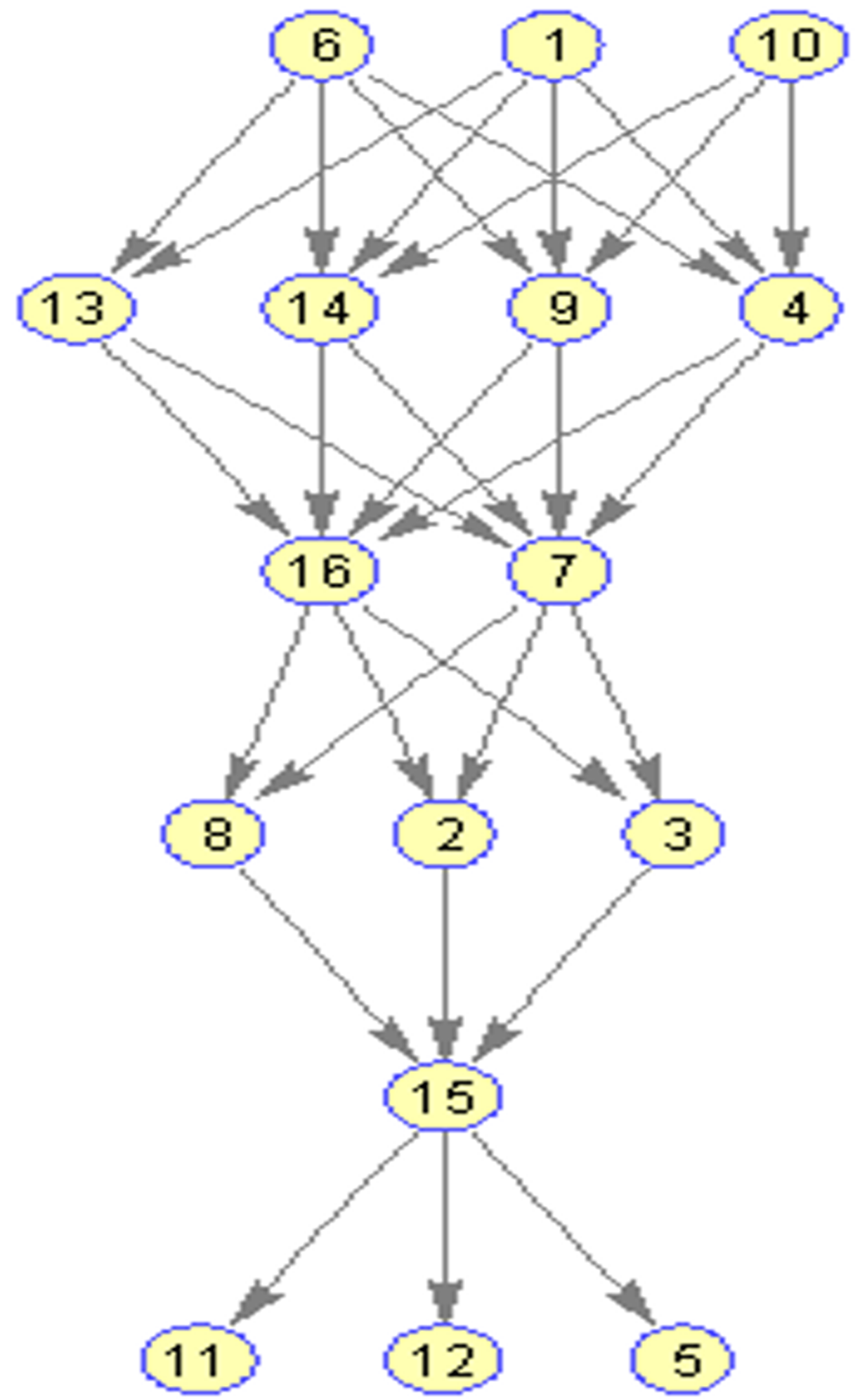}}
		\caption{Experimental results of Bradley-Terry model on IQA dataset.} \label{iqaranking}
	\end{center}
\end{figure}

\subsection{Image Quality Assessment}
\textbf{Dataset Description} The second dataset is for subjective image quality assessment (IQA), which contains 15 reference images and 15 distorted versions of each reference, for a total of 240 images which come from two publicly available datasets LIVE, \cite{LIVE} and IVC \cite{IVC}. Totally, 342 observers,
each of whom performs a varied number of comparisons via Internet, provide
$52,043$ feedbacks (i.e., 1, 0, -1) for crowdsourced subjective image quality assessment. For simplicity, we randomly take
reference 1 as an illustrative example while other reference
images exhibit similar results.\\
\textbf{Competitors} Now we introduce the competitors employed in our experiments. As mentioned in the Section \ref{sec:relatedwork}, the $\alpha$-cut algorithm shares the most similar problem setting with our proposed algorithm and thus is adopted as our main competitor. Seeing that the $\alpha$-cut algorithm employs bagging ensembles of weak learners, we further compare our proposed algorithms with  $\alpha$-cut algorithm when different types of such weak learners are adopted. \\
\textbf{Experiment Setting} Different from simulated data, as there are no ground-truth in real-world data, one can not compute Macro-F and Micro-F as in simulated data to evaluate the method we proposed. To see whether our proposed method could provide precise partial ranking, we generate 20 repetitions of training/testing splits with 80\% of the samples are selected as the training set and the rest as the testing set. Regarding the parameter-tuning of the weak learners in $\alpha$-cut, we tune the coefficient for Ridge/LASSO regularization from the range $\{2^{-7}, 2^{-6}, \cdots, 2^{-3}\}$ and the best parameter is selected via a 5-fold cross-validation on the training set.\\
\textbf{Evaluation metrics} To test whether the edges we added in the graph are reasonable or not, we employ two metrics called correctness and completeness, respectively. Given the true partial order relation $\succ_*$, the estimated partial order relation $\succ$, the concordant set:
\[\mathcal{C} = {\{(i,j):~ (i \succ j \wedge i ~{\succ}_{*}~ j) \vee (j\succ i \wedge  j \succ_{*} i)\}}\] and discordant set \[\mathcal{D}: ={\{(i,j):~ (i \succ j \wedge j ~{\succ}_{*}~ i) \vee (j\succ i \wedge  i \succ_{*} j)\}}\]
we could define a metric for completeness as :
\begin{equation*}
completeness = \frac{\left\vert \mathcal{C}\right\vert + \left\vert \mathcal{D}\right\vert}{\left\vert\{(i,j): i \succ_{*} j \vee j \succ_{*} i \}\right\vert}.
\end{equation*}
It is easy to find that the completeness metric measures the ability to detect a comparable pair. Likewise, correctness is defined as follows:
\begin{equation*}
correctness = \frac{\left\vert
	\mathcal{C}\right\vert}{\left\vert \mathcal{C}\right\vert + \left\vert \mathcal{D}\right\vert}.
\end{equation*}

{\renewcommand\baselinestretch{1.0}\selectfont
	
	\begin{table} [t]
 \renewcommand{\captionfont}{\footnotesize \bfseries}
		\caption{ Experimental results on IQA dataset.}
		
		\centering
		\small

		\begin{lrbox}{\tablebox}
			\begin{tabular}{ccccccccccc}
				\toprule
				\multirow{2}[4]{*}{types} & \multicolumn{1}{c}{\multirow{2}[4]{*}{algorithms}} &      & \multicolumn{2}{c}{correctness} &      & \multicolumn{2}{c}{completeness} &      & \multicolumn{2}{c}{geomean} \\
				\cmidrule{4-5}\cmidrule{7-8}\cmidrule{10-11}         &      &      & median & std  &      & median & std  &      & median & std \\
				\midrule
				\multirow{8}[1]{*}{$\alpha$-cut \cite{cheng2010predicting}} & LRLASSO  &      & 0.9137  & 0.0173  &      & 0.8309  & 0.0325  &      & 0.8760  & 0.0200  \\
				& LRRidge&      & 0.9227  & 0.0150  &      & 0.8044  & 0.0301  &      & 0.8582  & 0.0148  \\
				& SVMLASSO  &      & 0.9158  & 0.0137  &      & 0.8310  & 0.0297  &      & 0.8721  & 0.0166  \\
				& SVMRidge  &      & 0.9184  & {0.0099} &      & 0.8083  & 0.0484  &      & 0.8594  & 0.0246  \\
				& LSLASSO  &      & 0.9154  & 0.0117  &      & 0.8095  & 0.0285  &      & 0.8623  & 0.0146  \\
				& LSRidge  &      & 0.9139  & 0.0126  &      & 0.8218  & 0.0336  &      & 0.8668  & 0.0182  \\
				& SVRLASSO &      & {0.9236} & 0.0119  &      & 0.7405  & 0.0291  &      & 0.8311  & 0.0167  \\
				& SVRRidge &      & 0.9191  & 0.0145  &      & 0.7594  & 0.0386  &      & 0.8378  & 0.0187  \\
				\midrule
				\multirow{3}[1]{*}{ours} & \textbf{Uniform} &      & \textbf{\textbf{0.9137}}  & 0.0107  &      & \textbf{\textbf{0.8623}}  & 0.0142  &      &\textbf{\textbf{0.8867}} & 0.0081  \\
				& \textbf{Bradley-Terry} &      & \textbf{\textbf{0.9113}}  & 0.0124  &      & \textbf{\textbf{0.9254}}  & 0.0141  &      & \textbf{\textbf{0.9064}}  & 0.0082  \\& \textbf{Thurstone-Mosteller} &      & \textbf{\textbf{0.9146}}  & 0.0122  &      & \textbf{\textbf{0.9077}} & {0.0122} &      & \textbf{\textbf{0.9084}} & {0.0075} \\
				
				\bottomrule
			\end{tabular}
			
		\end{lrbox}
		\scalebox{0.65}{\usebox{\tablebox}}
		
		\label{tab:iqa_perf}%
		
	\end{table}
	\par}

According to the definition, we see that a higher correctness implies a more accurate prediction for the pairs which are detected as comparable.
Actually, there
is always a trade-off between these two criteria: correctness on the one side and completeness on the other
side. An ideal learner is correct in the sense of making few mistakes, but also complete in the sense
of abstaining rarely. In other words, the two criteria are conflicting: increasing completeness typically might as well come along
with reducing correctness and vice versa. Here we plot the trade-off between completeness and correctness as $\lambda$ varies. After all, every $\lambda$ can induce a partial ranking. The
partial ranking obtained by $\lambda$-cut of MLE is highlighted as red circle, as is shown in Figure \ref{iqaranking}(a). \\
\textbf{Performance Comparison} Table \ref{tab:iqa_perf} shows the corresponding performance of our proposed algorithms and the $\alpha$-cut algorithms. In this table, the second column shows the weak learner and regularization term employed in $\alpha$-cut and three models proposed in our algorithm. Specifically, LR represents for logistics regression \cite{prml}, SVM stands for the Support Vector Machine \cite{svm} method, LS stands for the method of least squares \cite{prml} while SVR stands for the Support Vector Regression \cite{svr} method. For regularization, we employ the Ridge \cite{ridge} and LASSO \cite{lasso} regularization term.  In order to comprehensively aggregate the performance, an overall metric should be defined based on both criteria. This leads to our inclusion of the last column which records the corresponding statistics for the geometric mean of the two mentioned criteria. According to this table, we find that our proposed algorithms significantly outperform other competitors in terms of completeness, and reach comparable results in terms of correctness. Moreover, the advantage in terms of the third metric also suggests the comprehensive superiority of our proposed algorithms.      \\
\textbf{Partial Order Visualization} Here Figure \ref{iqaranking}(b) depicts a diagram for the partial order induced by Bradley-Terry Model. Disconnected nodes in the diagram indicate
the incomparability of their corresponding subjective quality. Take the fourth level (ID=8,2,3) as an example. These three images come from
LIVE datasets \cite{LIVE}, and the corresponding names in LIVE dataset are ID=8 (img91-jp2k.bmp), ID=2 (img95-fastfading.bmp), ID=3 (img91-fastfading.bmp). Via our proposed partial ranking algorithm, the quality of three images are treated as confusing thus located on the same level. To see whether they are really confusing or not, we go back to check the mean opinion score (MOS) of three images provided by LIVE dataset. We are pleasantly surprised to find that their MOS are so close: 50.96, 50.29, 48.62, respectively. From this viewpoint, the partial ranking we obtained is reasonable. However, MOS is not always accurate enough,
which suffers from:
i) Unable to concretely define the concept of scale; ii) Dissimilar interpretations of the scale among users; iii) Difficult to verify whether a participant gives false ratings either intentionally or carelessly. In this case, the results derived from our method could stand up to undertake the mission of being the ground-truth for image quality assessment.

{\renewcommand\baselinestretch{1.0}\selectfont
	
	\begin{table} [t]
 \renewcommand{\captionfont}{\footnotesize \bfseries}
		\caption{ Experimental results on human age dataset.}
		
		\centering
		
		
		\small
		\begin{lrbox}{\tablebox}
			\begin{tabular}{cccccccccccc}
				\toprule
				\multirow{2}[4]{*}{type} & \multicolumn{1}{c}{\multirow{2}[4]{*}{algorithms}} &      & \multicolumn{2}{c}{correctness} &      & \multicolumn{2}{c}{completemess} &      & \multicolumn{2}{c}{geomean} \\
				\cmidrule{4-5}\cmidrule{7-8}\cmidrule{10-11}         &      &      & \multicolumn{1}{c}{median} & \multicolumn{1}{c}{std} &      & \multicolumn{1}{c}{median} & \multicolumn{1}{c}{std} &      & \multicolumn{1}{c}{median} & \multicolumn{1}{c}{std} \\
				\midrule
				\multirow{8}[1]{*}{$\alpha$-cut\cite{cheng2010predicting}} & LRLASSO  &      & 0.8640  & 0.0095  &      & 0.8352  & 0.0974  &      & 0.8511  & 0.0562  \\
				&LRRidge &      & 0.8693  & 0.0070  &      & 0.8467  & 0.0186  &      & 0.8584  & 0.0090  \\
				& SVMLASSO &      & 0.8674  & 0.0084  &      & 0.8565  & 0.0315  &      & 0.8619  & 0.0144  \\
				& SVMRidge  &      & 0.8660  & 0.0076  &      & 0.8447  & 0.1049  &      & 0.8542  & 0.0597  \\
				& LSLASSO   &      & 0.8688  & 0.0072  &      & 0.8583  & 0.0265  &      & 0.8617  & 0.0128  \\
				& LSRidge &      & 0.8681  & 0.0072  &      & 0.8513  & 0.0193  &      & 0.8556  & 0.0096  \\
				& SVRLASSO &      & {0.8732} & 0.0087  &      & 0.7687  & 0.0380  &      & 0.8177  & 0.0188  \\
				&SVRRidge &      & {0.8732} & 0.0082  &      & 0.7750  & 0.0229  &      & 0.8237  & 0.0118  \\
				\midrule
				\multirow{3}[1]{*}{ours} & \textbf{Uniform} &      &\textbf{\textbf{0.8655}}  & {0.0056} &      &\textbf{\textbf{0.8523}}  & 0.0098  &      &\textbf{\textbf{0.8591}}  & 0.0056  \\
				& \textbf{Bradley-Terry} &      &\textbf{\textbf{0.8671}}  & 0.0061  &      & \textbf{\textbf{0.8990}} & 0.0070  &      &\textbf{\textbf{0.8826}} & {0.0042} \\& \textbf{Thurstone-Mosteller} &      & \textbf{\textbf{0.8682}}  & 0.0062  &      &\textbf{\textbf{0.8949}}  & {0.0067} &      & \textbf{\textbf{0.8816}}  & 0.0044  \\

				\bottomrule
			\end{tabular}%
		\end{lrbox}
		\scalebox{0.65}{\usebox{\tablebox}}
		
		\label{tab:age_perf}%

	\end{table}
	\par}

\begin{figure}[t]
	 \renewcommand{\captionfont}{\footnotesize \bfseries}
	 \setlength{\belowcaptionskip}{-10pt}
	\begin{center}
		\includegraphics[width=0.4\textwidth]{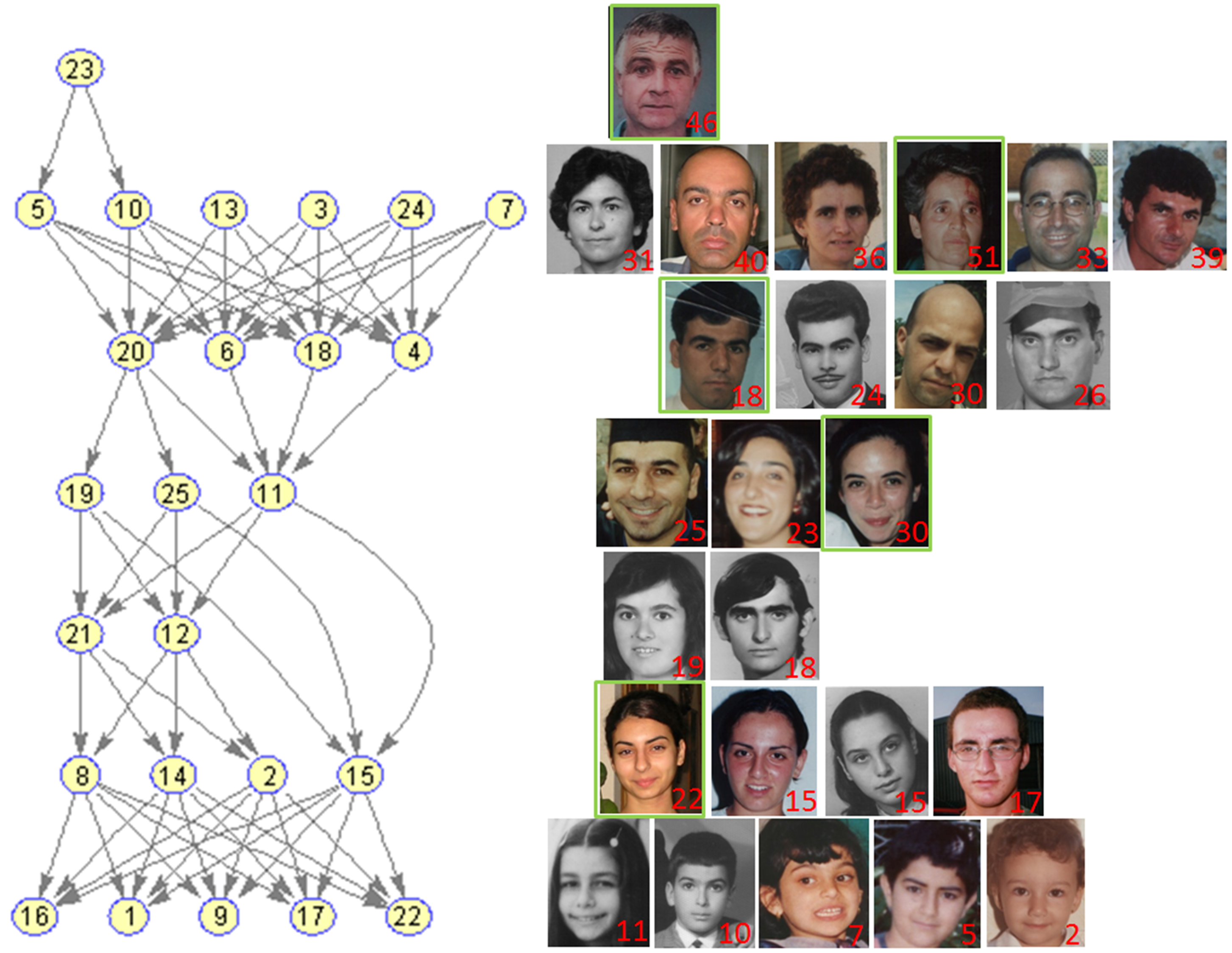}
		\caption{Partial ranking of human age dataset.} \label{ageranking}
	\end{center}
\end{figure}

\begin{figure}[t]
	 \renewcommand{\captionfont}{\footnotesize \bfseries}
	 \setlength{\belowcaptionskip}{-13pt}
	\setlength{\abovecaptionskip}{10pt}
	\begin{center}
		\subfigure[Conflict images]{
			\includegraphics[width=0.21\textwidth]{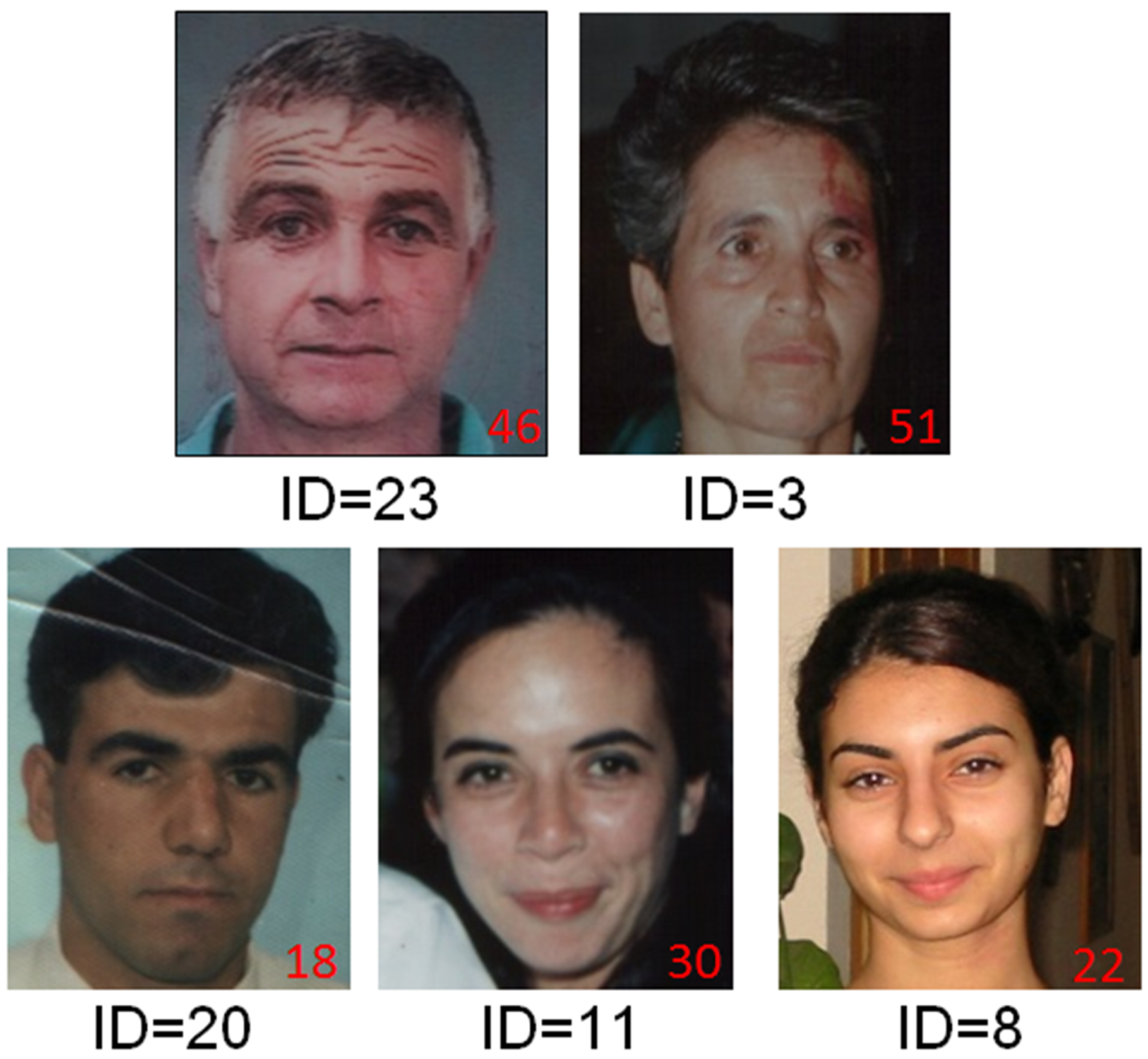}}
		\subfigure[Optimal $\lambda$]{
			\includegraphics[width=0.22\textwidth]{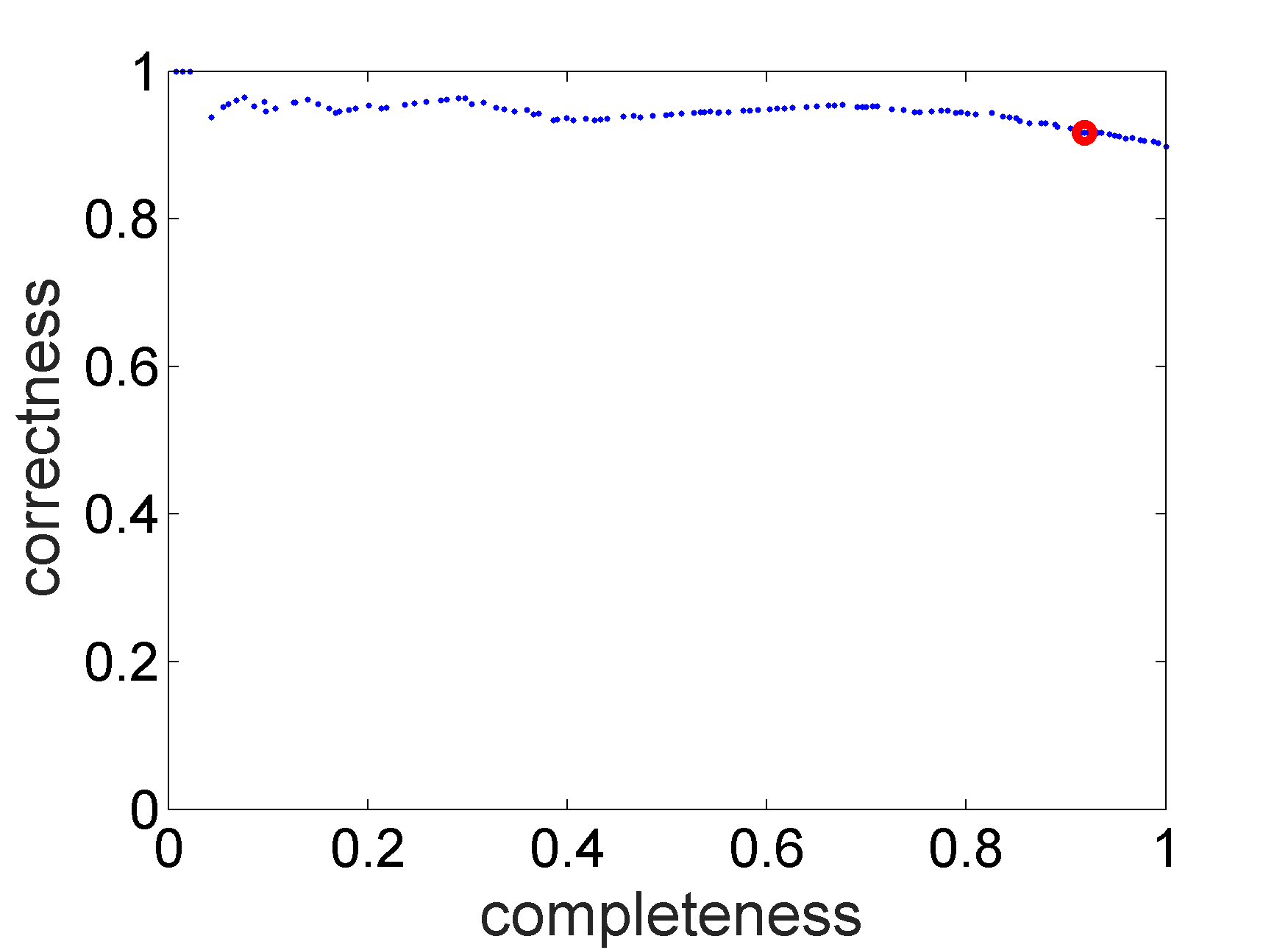}}
		\caption{Conflict images and optimal $\lambda$ of human age dataset.} \label{agesubset}
	\end{center}
\end{figure}

\subsection{Human Age}\label{sec:age}
\textbf{Dataset Description} In this dataset, 25 images from human age dataset FG-NET \footnote{http://www.fgnet.rsunit.com/} are annotated by a group of volunteer users on \href{http://www.chinacrowds.com/}{ChinaCrowds} platform. The groundtruth age ranking is known to us. The annotator is presented with two images and given a choice of which one is older (or difficult to judge). Totally, we obtain 9589 feedbacks from 91 annotators. \\
\textbf{Performance Comparison}
For age dataset, we adopt the same  experiment setting, competitors and
hyperparameter tuning strategy as the IQA dataset. Table \ref{tab:age_perf} shows the comparable results on this dataset. Similar with the results on the IQA dataset, we can
find that our proposed algorithms reach comparable performance in the sense of correctness. While, for the last two models (i.e. Bradley-Terry and Thurstone-Mosteller), our proposed algorithm significantly outperforms the competitors in terms of completeness. This leverages a better geometric mean of our algorithm with the last two models.\\
\textbf{Partial Ranking Visualization} Moreover, Figure \ref{ageranking} (Left) shows the partial ranking we obtained with 7 hierarchical levels on this dataset. It is easy to see that ID=23, the oldest, stands on the first level, while ID=5,10,13,3,24,7 are on the second level, and so on. On the leaf nodes, individuals with ID=16,1,9,17,22 are the youngest group of this dataset. To demonstrate whether the partial ranking we derived is reasonable or not, the original images are shown on the right panel, with ground truth ages painted red on the right corner of each image. From top to down, we can see that ID=23 (46 years old) is indeed older than most of the individuals on level 2 except ID=3 (51 years old). In other words, the partial ranking by mistake thinks 46 older than 51! If we look into the details of these two individuals, as is shown in Figure \ref{agesubset}(a), the man with ID=23 gets more wrinkles, especially around his forehead and eyes, compared with the woman with ID=3. Besides, the man has white hair on his temples while the woman not. Another three conflicts happen on level 3 of ID=20 (18 years old), level 4 of ID=11 (30 years old), and level 6 of ID=8 (22 years old), respectively. We guess the reason behind lies in the three individuals have more or less the gap with their actual ages. For example, ID=20 looks older than he really is, while other two women (ID=11 and ID=8) look significantly younger than they really are. Especially the woman ID=8 with 22 years old looks even younger than other two girls (ID=14 and ID=2) who are actually 7 years younger than her. From this viewpoint, the partial ranking derived from our proposed method is reasonable. Moreover, the optimal $\lambda$ on this dataset is highlighted as red circle, as is shown in Figure \ref{agesubset}(b).

{\renewcommand\baselinestretch{1.0}\selectfont
	
	\begin{table} [t]
		\renewcommand{\captionfont}{\footnotesize \bfseries}
		\setlength{\belowcaptionskip}{10pt}
		\setlength{\abovecaptionskip}{5pt}
		\caption{ Experimental results on worldCollege dataset.}
		\small
		\centering
		\begin{lrbox}{\tablebox}
			\begin{tabular}{cccccccccccc}
				\toprule
				\multirow{2}[4]{*}{types} & \multicolumn{1}{c}{\multirow{2}[4]{*}{algorithms}} &      & \multicolumn{2}{c}{correctness} &      & \multicolumn{2}{c}{completeness} &      & \multicolumn{2}{c}{geomean} \\
				\cmidrule{4-5}\cmidrule{7-8}\cmidrule{10-11}         &      &      & \multicolumn{1}{c}{median} & \multicolumn{1}{c}{std} &      & \multicolumn{1}{c}{median} & \multicolumn{1}{c}{std} &      & \multicolumn{1}{c}{median} & \multicolumn{1}{c}{std} \\
				\midrule
				\multirow{8}[2]{*}{$\alpha$-cut \cite{cheng2010predicting} } & LRLASSO &      & 0.5100  & 0.0121  &      & 1.0000  & 0.0412  &      & 0.7135  & 0.0120  \\
				& LRRidge &      & 0.7439  & 0.0139  &      & 0.7475  & 0.0639  &      & 0.7391  & 0.0285  \\
				& SVMLASSO  &      & 0.5090  & 0.0091 &      & 1.0000  & 0.0012  &      & 0.7135  & 0.0063  \\
				& SVMRidge &      & 0.7488  & 0.0150  &      & 0.7531  & 0.0655  &      & 0.7448  & 0.0297  \\
				& LSLASSO &      & 0.5090  & 0.0091&      & 1.0000  & 0.0000  &      & 0.7135  & 0.0064  \\
				& LSRidge &      & 0.7490  & 0.0117  &      & 0.6518  & 0.0699  &      & 0.7020  & 0.0325  \\
				& SVRLASSO &      & 0.5090  & 0.0091 &      & 1.0000  & 0.0000  &      & 0.7135  & 0.0064  \\
				& SVRRidge &      & 0.7463  & 0.0151  &      & 0.7394  & 0.0671  &      & 0.7373  & 0.0300  \\
				\midrule
				\multirow{3}[2]{*}{ours} & \textbf{Uniform }&      & \textbf{\textbf{0.7557}}  & 0.0101  &      & \textbf{\textbf{0.7478}}  & 0.0104  &      & \textbf{\textbf{0.7501}}  & 0.0074  \\
				& \textbf{Bradley-Terry} &      & \textbf{\textbf{0.7629}} & 0.0108  &      & \textbf{\textbf{0.7566}}  & 0.0087  &      & \textbf{\textbf{0.7583}} & 0.0069  \\
				& \textbf{Thurstone-Mosteller}&      & \textbf{\textbf{0.7619}}  & 0.0110  &      & \textbf{\textbf{0.7586}} & 0.0082 &      & \textbf{\textbf{0.7576}}  & 0.0066  \\
				
				\bottomrule
			\end{tabular}%
		\end{lrbox}
		\scalebox{0.68}{\usebox{\tablebox}}
		\label{tab:college_perf}
		
	\end{table}
	\par}

\subsection{WorldCollege Ranking}
\begin{figure}[t]
 \renewcommand{\captionfont}{\footnotesize \bfseries}
	\setlength{\belowcaptionskip}{-5pt}
	\setlength{\abovecaptionskip}{5pt}
	\begin{center}
		\subfigure[Optimal $\lambda$]{
			\includegraphics[width=0.23\textwidth]{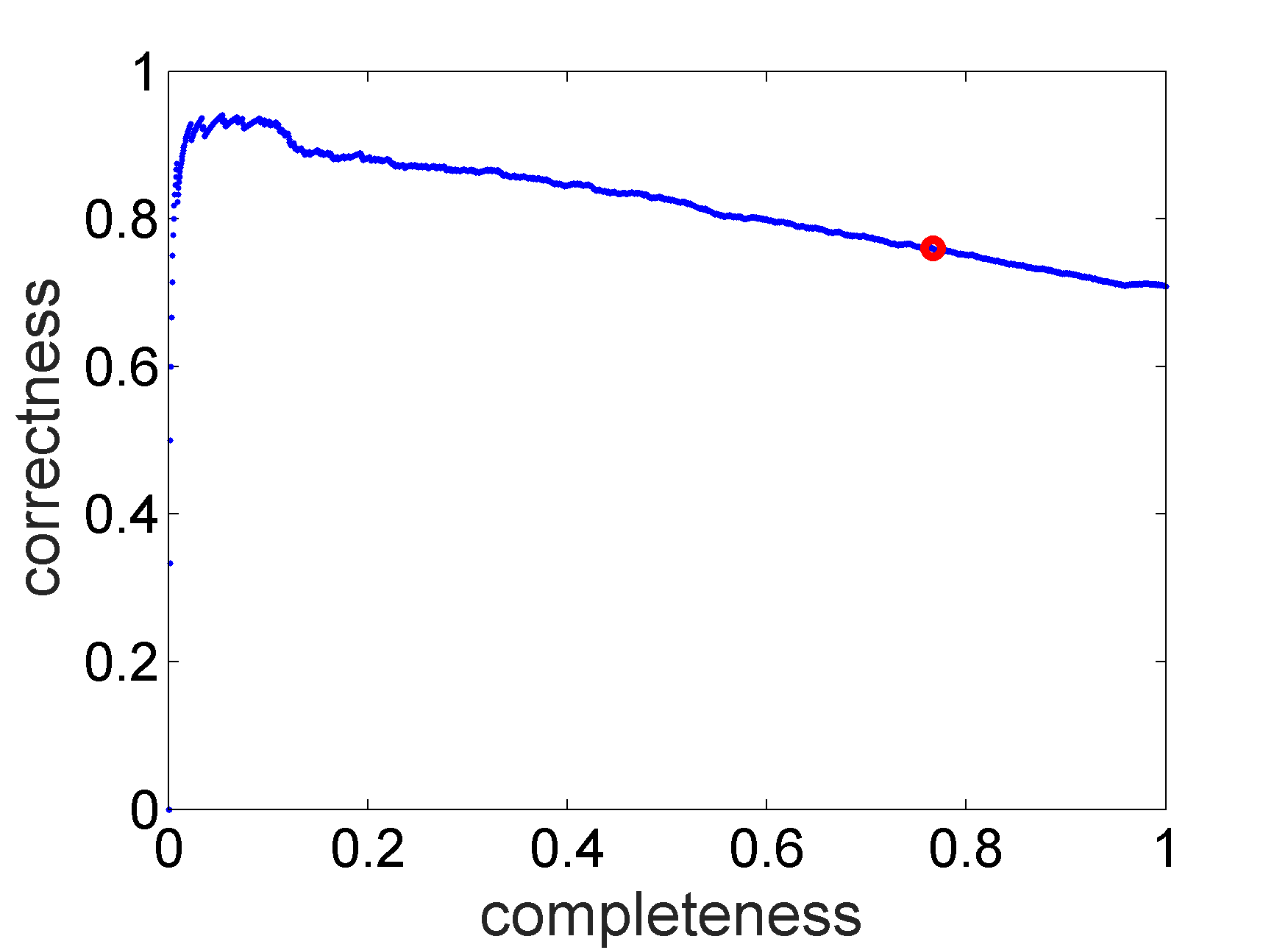}}
		\subfigure[World map]{
			\includegraphics[width=0.23\textwidth]{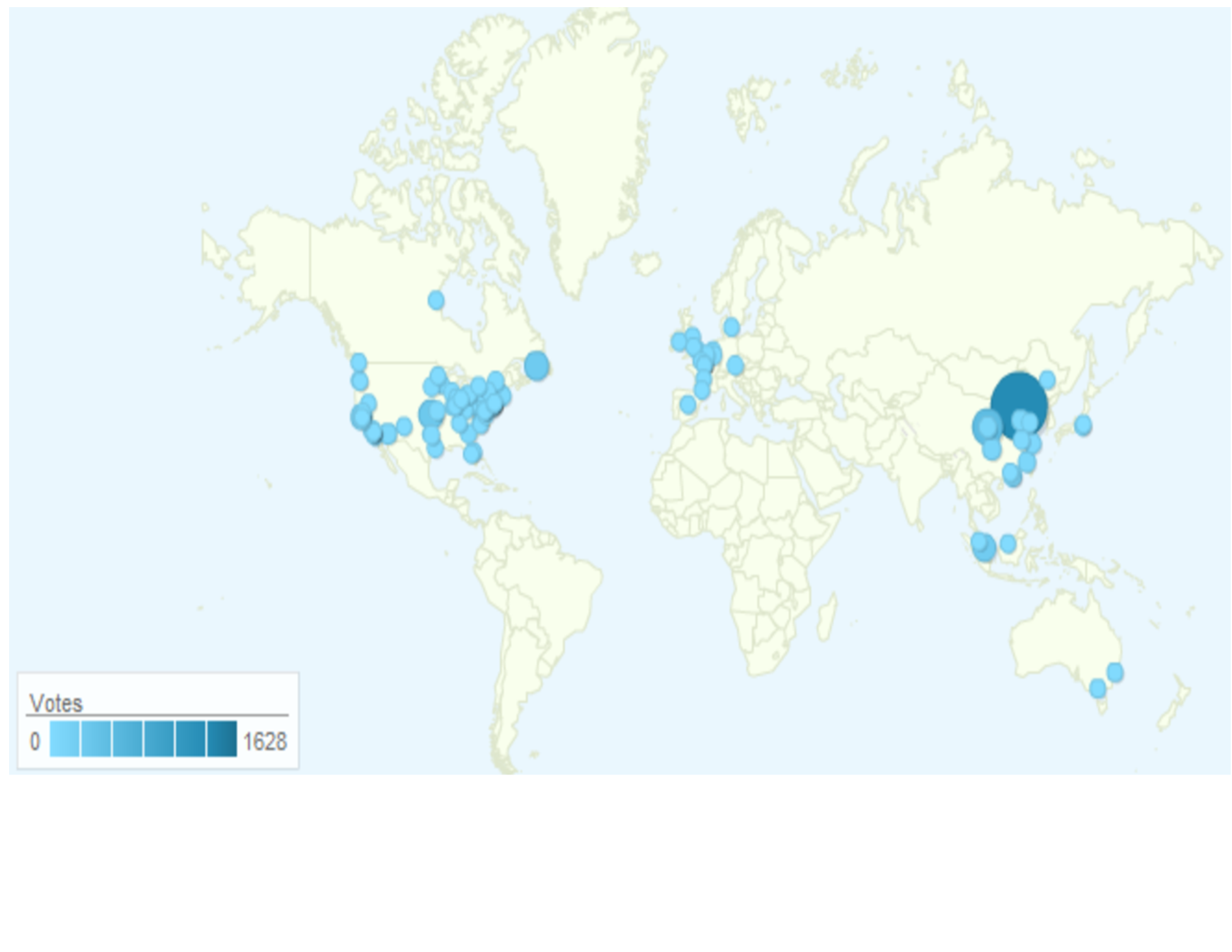}}
		\caption{Optimal $\lambda$ and world map of all the annotators in worldCollege dataset.} \label{optimalcollege}
	\end{center}
\end{figure}

\begin{figure}[t]
 \renewcommand{\captionfont}{\footnotesize \bfseries}
	 \setlength{\belowcaptionskip}{-5pt}
	\setlength{\abovecaptionskip}{5pt}
	\begin{center}
		\includegraphics[width=0.48\textwidth]{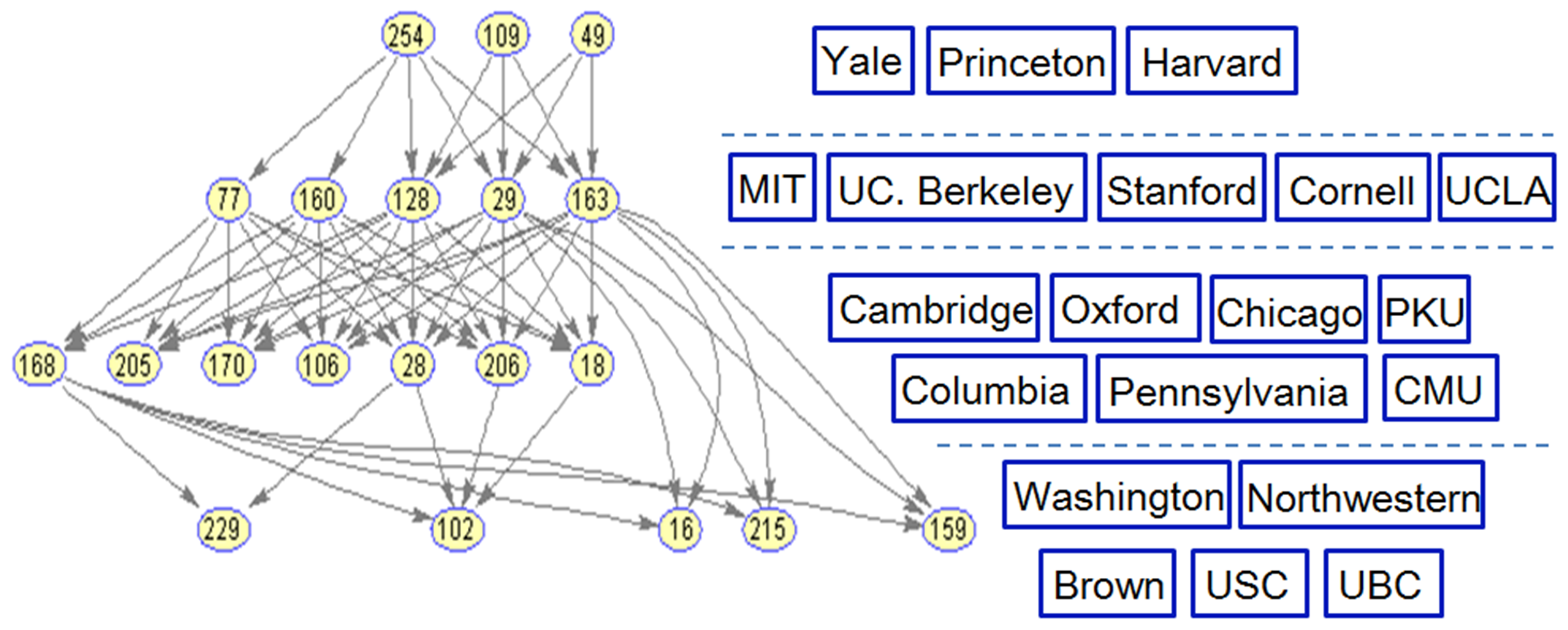}
		\caption{Partial ranking of worldCollege dataset.} \label{collegeranking}
	\end{center}
\end{figure}

\textbf{Data Description} We now apply the proposed method to the worldCollege ranking dataset, which is composed of 261 colleges. Using the \href{http://www.allourideas.org/}{Allourideas} crowdsourcing platform, a total of 340 distinct annotators from various countries (e.g., USA, Canada, Spain, France, Japan, China, etc.) are shown randomly with pairs of these colleges, and asked to decide which of the two universities is more attractive to attend. If the voter thinks the two colleges are incomparable, he can choose the third option by clicking ``I can't decide". Finally, we obtain a total of 11012 feedbacks, among which 9409 samples are pairwise comparisons with clear opinions and the remaining 1603 are records with voter clicking ``I can't decide". \\
\textbf{Performance Comparisons} Table \ref{tab:college_perf} shows the comparable results on the college dataset. It is easy to see that our proposed algorithms again attain better correctness than all the $\alpha$-cut variants. Moreover, we find that all the LASSO-based $\alpha$-cut variants exhibit almost perfect completeness. Nonetheless, this superiority on completeness comes at a fatal price: the corresponding correctness results are close to 0.5, a value for a random ranker. Having perfect completeness alone thus does not make LASSO variants the top rankers. Consequently, in view of the aggregated metric, we see that all the LASSO variants show unreasonable performance on the third column while our proposed algorithms attain better comprehensive performance than $\alpha$-cut variants. Furthermore, compared to the two real-world datasets above, the performance on this dataset is a little bit worse than the IQA and human age datasets. We then go back to the
crowdsourcing platform and find out that the reason
behind lies in the ``I can't decide" button. Though most voters click this button when he thinks two colleges are incomparable and difficult to choose, there are also some voters click this button because he does not know both of these two colleges or one of them. From this viewpoint, colleges with distinguishable difference even have the possibility to be treated as incomparable, just because the voters are not familiar with them.
Due to the existence of these contaminated samples, though the performance of our proposed method declines by 10\% approximately on this dataset, we still think it a reasonable phenomenon. Besides, the optimal $\lambda$ on this dataset is illustrated in Figure \ref{optimalcollege}(a). \\
\textbf{Partial Order Visualization} Considering the partial ranking on 261 colleges is difficult to show, we only illustrate the partial ranking on top-20 colleges in Figure \ref{collegeranking}. It is easy to see that
Yale, Princeton, and Harvard are the top 3 at the first level, while MIT, UC. Berkeley, Stanford, Cornell, UCLA are the second level. These results we derived are basically matched with the college ranking in reality.
But a mystery has emerged from the experimental results. That is, Peking University (PKU) magically jumped into the third echelon together with Cambridge, Oxford, CMU, etc. To investigate the reason behind this phenomenon, we go back to see the world map of all the
annotators. As is shown in Figure \ref{optimalcollege}(b), most of the annotators come from China, thus significantly raises the ranking of PKU located in the capital of China---Beijing.

\section{Conclusions}\label{sec:conclusions}
In this paper, we propose a partial ranking algorithm based on margin-based MLE to learn partial but more accurate (i.e., making less incorrect
statements) orders in crowdsourced ranking. In this scheme, three kinds of models are systematically discussed, including the uniform model, the Bradley-Terry model, and the Thurstone-Mosteller model. Moreover, we conduct theoretical analysis on FDR and Power control to demonstrate the
effectiveness of the proposed method.
Experimental studies conducted on simulated examples and three real-world
datasets show that our proposed method could exhibit better performance compared with the traditional methods. Our results suggest that the proposed methodology is an effective tool to provide partial ranking
for modern crowdsourced preference data.

\section{Acknowledgments}

The research of Qianqian Xu was supported in part by National Key R\&D Program of China (Grant No. 2016YFB0800403), National Natural Science Foundation of China (No.61672514, 61390514, 61572042), Beijing Natural Science Foundation (4182079), Youth Innovation Promotion Association CAS, and CCF-Tencent Open Research Fund. The research of Zhiyong Yang and Qingming Huang was supported in part by National Natural Science Foundation of China: 61332016, 61620106009, U1636214 and 61650202, in part by National Basic Research Program of China (973 Program): 2015CB351800, in part by Key Research Program of Frontier Sciences, CAS: QYZDJ-SSW-SYS013. The research of Yuan Yao was supported in part by Hong Kong Research Grant Council (HKRGC) grant 16303817, National Basic Research Program of
China (No. 2015CB85600, 2012CB825501), National Natural Science Foundation of China (No. 61370004, 11421-110001), as well as awards
from Tencent AI Lab, Si Family Foundation, Baidu Big Data Institute, and Microsoft Research-Asia.

\bibliographystyle{ACM-Reference-Format}

\bibliography{sigproc}  


\begin{thebibliography}{39}


\ifx \showCODEN    \undefined \def \showCODEN     #1{\unskip}     \fi
\ifx \showDOI      \undefined \def \showDOI       #1{#1}\fi
\ifx \showISBNx    \undefined \def \showISBNx     #1{\unskip}     \fi
\ifx \showISBNxiii \undefined \def \showISBNxiii  #1{\unskip}     \fi
\ifx \showISSN     \undefined \def \showISSN      #1{\unskip}     \fi
\ifx \showLCCN     \undefined \def \showLCCN      #1{\unskip}     \fi
\ifx \shownote     \undefined \def \shownote      #1{#1}          \fi
\ifx \showarticletitle \undefined \def \showarticletitle #1{#1}   \fi
\ifx \showURL      \undefined \def \showURL       {\relax}        \fi
\providecommand\bibfield[2]{#2}
\providecommand\bibinfo[2]{#2}
\providecommand\natexlab[1]{#1}
\providecommand\showeprint[2][]{arXiv:#2}

\bibitem[\protect\citeauthoryear{Arrow}{Arrow}{1963}]%
        {Arrow51}
\bibfield{author}{\bibinfo{person}{K. Arrow}.} \bibinfo{year}{1963}\natexlab{}.
\newblock \bibinfo{booktitle}{\emph{Social Choice and Individual Values, 2nd
  Ed.}}
\newblock \bibinfo{publisher}{Yale University Press, New Haven, CT}.
\newblock


\bibitem[\protect\citeauthoryear{Brin and Page}{Brin and Page}{1998}]%
        {Pagerank}
\bibfield{author}{\bibinfo{person}{Sergey Brin} {and} \bibinfo{person}{Larry
  Page}.} \bibinfo{year}{1998}\natexlab{}.
\newblock \showarticletitle{The Anatomy of a Large-Scale Hypertextual Web
  Search Engine}. In \bibinfo{booktitle}{\emph{International Conference on
  World Wide Web}}. \bibinfo{pages}{107--117}.
\newblock


\bibitem[\protect\citeauthoryear{Chen, Wu, Chang, and Lei}{Chen
  et~al\mbox{.}}{2009}]%
        {MM09}
\bibfield{author}{\bibinfo{person}{Kuan{-}Ta Chen}, \bibinfo{person}{Chen{-}Chi
  Wu}, \bibinfo{person}{Yu{-}Chun Chang}, {and} \bibinfo{person}{Chin{-}Laung
  Lei}.} \bibinfo{year}{2009}\natexlab{}.
\newblock \showarticletitle{A crowdsourceable QoE evaluation framework for
  multimedia content}. In \bibinfo{booktitle}{\emph{ACM International
  Conference on Multimedia}}. \bibinfo{pages}{491--500}.
\newblock


\bibitem[\protect\citeauthoryear{Cheng, Rademaker, De~Baets, and
  H{\"u}llermeier}{Cheng et~al\mbox{.}}{2010}]%
        {cheng2010predicting}
\bibfield{author}{\bibinfo{person}{Weiwei Cheng}, \bibinfo{person}{Micha{\"e}l
  Rademaker}, \bibinfo{person}{Bernard De~Baets}, {and} \bibinfo{person}{Eyke
  H{\"u}llermeier}.} \bibinfo{year}{2010}\natexlab{}.
\newblock \showarticletitle{Predicting partial orders: ranking with
  abstention}.
\newblock \bibinfo{journal}{\emph{Machine Learning and Knowledge Discovery in
  Databases}} (\bibinfo{year}{2010}), \bibinfo{pages}{215--230}.
\newblock


\bibitem[\protect\citeauthoryear{Chow}{Chow}{1970}]%
        {Chow1970}
\bibfield{author}{\bibinfo{person}{C. Chow}.} \bibinfo{year}{1970}\natexlab{}.
\newblock \showarticletitle{On optimum recognition error and reject tradeoff}.
\newblock \bibinfo{journal}{\emph{IEEE Transactions on Information Theory}}
  \bibinfo{volume}{16}, \bibinfo{number}{1} (\bibinfo{year}{1970}),
  \bibinfo{pages}{41--46}.
\newblock


\bibitem[\protect\citeauthoryear{Christopher}{Christopher}{2006}]%
        {prml}
\bibfield{author}{\bibinfo{person}{M~Bishop Christopher}.}
  \bibinfo{year}{2006}\natexlab{}.
\newblock \bibinfo{booktitle}{\emph{Pattern Recognition and Machine Learning}}.
\newblock \bibinfo{publisher}{Springer-Verlag New York}.
\newblock


\bibitem[\protect\citeauthoryear{Cortes, Mohri, and Rastogi}{Cortes
  et~al\mbox{.}}{2007}]%
        {CorMohRas07}
\bibfield{author}{\bibinfo{person}{C. Cortes}, \bibinfo{person}{M. Mohri},
  {and} \bibinfo{person}{A. Rastogi}.} \bibinfo{year}{2007}\natexlab{}.
\newblock \showarticletitle{Magnitude-preserving ranking algorithms}. In
  \bibinfo{booktitle}{\emph{International Conference on Machine learning}}.
  \bibinfo{pages}{169--176}.
\newblock


\bibitem[\protect\citeauthoryear{Cristianini and Shawe-Taylor}{Cristianini and
  Shawe-Taylor}{2000}]%
        {svm}
\bibfield{author}{\bibinfo{person}{N. Cristianini} {and} \bibinfo{person}{J.
  Shawe-Taylor}.} \bibinfo{year}{2000}\natexlab{}.
\newblock \bibinfo{booktitle}{\emph{An Introduction to Support Vector Machines
  and Other Kernel-based Learning Methods}}.
\newblock \bibinfo{publisher}{Cambridge Unversity Press}.
\newblock


\bibitem[\protect\citeauthoryear{Cynthia, Ravi, Moni, and Dandapani}{Cynthia
  et~al\mbox{.}}{2001}]%
        {dwork2001rank}
\bibfield{author}{\bibinfo{person}{D. Cynthia}, \bibinfo{person}{K. Ravi},
  \bibinfo{person}{N. Moni}, {and} \bibinfo{person}{S. Dandapani}.}
  \bibinfo{year}{2001}\natexlab{}.
\newblock \showarticletitle{Rank aggregation methods for the web}. In
  \bibinfo{booktitle}{\emph{International Conference on World Wide Web}}.
  \bibinfo{pages}{613--622}.
\newblock


\bibitem[\protect\citeauthoryear{David}{David}{1988}]%
        {David88}
\bibfield{author}{\bibinfo{person}{H. David}.} \bibinfo{year}{1988}\natexlab{}.
\newblock \bibinfo{booktitle}{\emph{The Method of Paired Comparisons}}.
\newblock \bibinfo{publisher}{Oxford University Press, New York, NY}.
\newblock


\bibitem[\protect\citeauthoryear{de~Borda}{de~Borda}{1781}]%
        {de1781memoire}
\bibfield{author}{\bibinfo{person}{J. de Borda}.}
  \bibinfo{year}{1781}\natexlab{}.
\newblock \bibinfo{booktitle}{\emph{M{\'e}moire sur les Elections au Scrutin}}.
\newblock \bibinfo{publisher}{Histoire de l'Acad{\'e}mie Royale des Sciences}.
\newblock


\bibitem[\protect\citeauthoryear{de~Condorcet}{de~Condorcet}{1785}]%
        {Condorcet}
\bibfield{author}{\bibinfo{person}{Marquis de Condorcet}.}
  \bibinfo{year}{1785}\natexlab{}.
\newblock \bibinfo{booktitle}{\emph{\'{E}ssai sur l'Application de l'Analyse
  \`{a} la Probabilit\'{e} des D\'{e}cisions Rendues \`{a} la Pluralit\'{e} des
  Voix (Essay on the Application of Analysis to the Probability of Majority
  Decisions)}}.
\newblock \bibinfo{publisher}{Imprimerie Royale, Paris}.
\newblock


\bibitem[\protect\citeauthoryear{Drucker, Burges, Kaufman, Smola, and
  Vapnik}{Drucker et~al\mbox{.}}{1996}]%
        {svr}
\bibfield{author}{\bibinfo{person}{Harris Drucker},
  \bibinfo{person}{Christopher J.~C. Burges}, \bibinfo{person}{Linda Kaufman},
  \bibinfo{person}{Alexander~J. Smola}, {and} \bibinfo{person}{Vladimir
  Vapnik}.} \bibinfo{year}{1996}\natexlab{}.
\newblock \showarticletitle{Support Vector Regression Machines}. In
  \bibinfo{booktitle}{\emph{Advances in Neural Information Processing
  Systems}}. \bibinfo{pages}{155--161}.
\newblock


\bibitem[\protect\citeauthoryear{Gionis, Mannila, Puolam{\"a}ki, and
  Ukkonen}{Gionis et~al\mbox{.}}{2006}]%
        {gionis2006algorithms}
\bibfield{author}{\bibinfo{person}{Aristides Gionis}, \bibinfo{person}{Heikki
  Mannila}, \bibinfo{person}{Kai Puolam{\"a}ki}, {and} \bibinfo{person}{Antti
  Ukkonen}.} \bibinfo{year}{2006}\natexlab{}.
\newblock \showarticletitle{Algorithms for discovering bucket orders from
  data}. In \bibinfo{booktitle}{\emph{ACM SIGKDD International Conference on
  Knowledge Discovery and Data Mining}}. \bibinfo{pages}{561--566}.
\newblock


\bibitem[\protect\citeauthoryear{Hoerl and Kennard}{Hoerl and Kennard}{1970}]%
        {ridge}
\bibfield{author}{\bibinfo{person}{Arthur~E Hoerl} {and}
  \bibinfo{person}{Robert~W Kennard}.} \bibinfo{year}{1970}\natexlab{}.
\newblock \showarticletitle{Ridge regression: Biased estimation for
  nonorthogonal problems}.
\newblock \bibinfo{journal}{\emph{Technometrics}} \bibinfo{volume}{12},
  \bibinfo{number}{1} (\bibinfo{year}{1970}), \bibinfo{pages}{55--67}.
\newblock


\bibitem[\protect\citeauthoryear{Jiang, Lim, Yao, and Ye.}{Jiang
  et~al\mbox{.}}{2011}]%
        {hodge}
\bibfield{author}{\bibinfo{person}{X. Jiang}, \bibinfo{person}{L-H. Lim},
  \bibinfo{person}{Y. Yao}, {and} \bibinfo{person}{Y. Ye.}}
  \bibinfo{year}{2011}\natexlab{}.
\newblock \showarticletitle{Statistical ranking and combinatorial {H}odge
  theory}.
\newblock \bibinfo{journal}{\emph{Mathematical Programming}}
  \bibinfo{volume}{127}, \bibinfo{number}{6} (\bibinfo{year}{2011}),
  \bibinfo{pages}{203--244}.
\newblock


\bibitem[\protect\citeauthoryear{Kleinberg}{Kleinberg}{1999}]%
        {Hits}
\bibfield{author}{\bibinfo{person}{Jon Kleinberg}.}
  \bibinfo{year}{1999}\natexlab{}.
\newblock \showarticletitle{Authoritative sources in a hyperlinked
  environment}.
\newblock \bibinfo{journal}{\emph{J. ACM}} \bibinfo{volume}{46},
  \bibinfo{number}{5} (\bibinfo{year}{1999}), \bibinfo{pages}{604--632}.
\newblock


\bibitem[\protect\citeauthoryear{Le~Callet and Autrusseau}{Le~Callet and
  Autrusseau}{2005}]%
        {IVC}
\bibfield{author}{\bibinfo{person}{Patrick Le~Callet} {and}
  \bibinfo{person}{Florent Autrusseau}.} \bibinfo{year}{2005}\natexlab{}.
\newblock \bibinfo{title}{Subjective quality assessment IRCCyN/IVC database}.
\newblock   (\bibinfo{year}{2005}).
\newblock
\newblock
\shownote{\url{http://www.irccyn.ec-nantes.fr/ivcdb/}.}


\bibitem[\protect\citeauthoryear{Lebanon and Mao}{Lebanon and Mao}{2008}]%
        {lebanon2008non}
\bibfield{author}{\bibinfo{person}{Guy Lebanon} {and} \bibinfo{person}{Yi
  Mao}.} \bibinfo{year}{2008}\natexlab{}.
\newblock \showarticletitle{Non-parametric modeling of partially ranked data}.
\newblock \bibinfo{journal}{\emph{Journal of Machine Learning Research}}
  \bibinfo{volume}{9} (\bibinfo{year}{2008}), \bibinfo{pages}{2401--2429}.
\newblock


\bibitem[\protect\citeauthoryear{Ma, Morel, Osher, and Chien}{Ma
  et~al\mbox{.}}{2011}]%
        {Osher11_retinex}
\bibfield{author}{\bibinfo{person}{W. Ma}, \bibinfo{person}{J.~M. Morel},
  \bibinfo{person}{S. Osher}, {and} \bibinfo{person}{A. Chien}.}
  \bibinfo{year}{2011}\natexlab{}.
\newblock \showarticletitle{An ${L}_1$-based variational model for Retinex
  theory and its application to medical images}. In
  \bibinfo{booktitle}{\emph{IEEE Conference on Computer Vision and Pattern
  Recognition}}. \bibinfo{pages}{153--160}.
\newblock


\bibitem[\protect\citeauthoryear{Negahban, Oh, and Shah}{Negahban
  et~al\mbox{.}}{2012}]%
        {negahban2012}
\bibfield{author}{\bibinfo{person}{S. Negahban}, \bibinfo{person}{S. Oh}, {and}
  \bibinfo{person}{D. Shah}.} \bibinfo{year}{2012}\natexlab{}.
\newblock \showarticletitle{Iterative ranking from pair-wise comparisons}. In
  \bibinfo{booktitle}{\emph{Annual Conference on Neural Information Processing
  Systems}}. \bibinfo{pages}{2483--2491}.
\newblock


\bibitem[\protect\citeauthoryear{Noether}{Noether}{1960}]%
        {Noether60}
\bibfield{author}{\bibinfo{person}{G. Noether}.}
  \bibinfo{year}{1960}\natexlab{}.
\newblock \showarticletitle{Remarks about a paired comparison model}.
\newblock \bibinfo{journal}{\emph{Psychometrika}}  \bibinfo{volume}{25}
  (\bibinfo{year}{1960}), \bibinfo{pages}{357--367}.
\newblock


\bibitem[\protect\citeauthoryear{Osting, Darbon, and Osher}{Osting
  et~al\mbox{.}}{2013}]%
        {osting2013statistical}
\bibfield{author}{\bibinfo{person}{Braxton Osting},
  \bibinfo{person}{J{\'e}r{\^o}me Darbon}, {and} \bibinfo{person}{Stanley
  Osher}.} \bibinfo{year}{2013}\natexlab{}.
\newblock \showarticletitle{STATISTICAL RANKING USING THE $L_1$-NORM ON
  GRAPHS.}
\newblock \bibinfo{journal}{\emph{Inverse Problems and Imaging}}
  \bibinfo{volume}{7}, \bibinfo{number}{3} (\bibinfo{year}{2013}),
  \bibinfo{pages}{907--926}.
\newblock


\bibitem[\protect\citeauthoryear{Parikh and Grauman}{Parikh and
  Grauman}{2011}]%
        {parikh2011relative}
\bibfield{author}{\bibinfo{person}{Devi Parikh} {and} \bibinfo{person}{Kristen
  Grauman}.} \bibinfo{year}{2011}\natexlab{}.
\newblock \showarticletitle{Relative attributes}. In
  \bibinfo{booktitle}{\emph{IEEE International Conference on Computer Vision}}.
  \bibinfo{pages}{503--510}.
\newblock


\bibitem[\protect\citeauthoryear{Rajkumar and Agarwal}{Rajkumar and
  Agarwal}{2014}]%
        {ICML14}
\bibfield{author}{\bibinfo{person}{Arun Rajkumar} {and}
  \bibinfo{person}{Shivani Agarwal}.} \bibinfo{year}{2014}\natexlab{}.
\newblock \showarticletitle{A Statistical Convergence Perspective of Algorithms
  for Rank Aggregation from Pairwise Data}. In
  \bibinfo{booktitle}{\emph{International Conference on Machine Learning}}.
  \bibinfo{pages}{118--126}.
\newblock


\bibitem[\protect\citeauthoryear{Saaty}{Saaty}{1977}]%
        {Saaty77}
\bibfield{author}{\bibinfo{person}{T. Saaty}.} \bibinfo{year}{1977}\natexlab{}.
\newblock \showarticletitle{A scaling method for priorities in hierarchical
  structures}.
\newblock \bibinfo{journal}{\emph{Journal of Mathematical Psychology}}
  \bibinfo{volume}{15}, \bibinfo{number}{3} (\bibinfo{year}{1977}),
  \bibinfo{pages}{234--281}.
\newblock


\bibitem[\protect\citeauthoryear{Saaty and Ozdemir}{Saaty and Ozdemir}{2003}]%
        {saaty2003magic}
\bibfield{author}{\bibinfo{person}{Thomas~L Saaty} {and}
  \bibinfo{person}{Mujgan~S Ozdemir}.} \bibinfo{year}{2003}\natexlab{}.
\newblock \showarticletitle{Why the magic number seven plus or minus two}.
\newblock \bibinfo{journal}{\emph{Mathematical and computer modelling}}
  \bibinfo{volume}{38}, \bibinfo{number}{3-4} (\bibinfo{year}{2003}),
  \bibinfo{pages}{233--244}.
\newblock


\bibitem[\protect\citeauthoryear{Schervish}{Schervish}{2012}]%
        {asymp}
\bibfield{author}{\bibinfo{person}{Mark~J Schervish}.}
  \bibinfo{year}{2012}\natexlab{}.
\newblock \bibinfo{booktitle}{\emph{Theory of Statistics}}.
\newblock \bibinfo{publisher}{Springer Science \& Business Media}.
\newblock


\bibitem[\protect\citeauthoryear{Sheikh, Z.Wang, Cormack, and Bovik}{Sheikh
  et~al\mbox{.}}{2008}]%
        {LIVE}
\bibfield{author}{\bibinfo{person}{H.R. Sheikh}, \bibinfo{person}{Z.Wang},
  \bibinfo{person}{L. Cormack}, {and} \bibinfo{person}{A.C. Bovik}.}
  \bibinfo{year}{2008}\natexlab{}.
\newblock \bibinfo{title}{{L}{I}{V}{E} Image \& Video Quality Assessment
  Database}.
\newblock   (\bibinfo{year}{2008}).
\newblock


\bibitem[\protect\citeauthoryear{Sismanis}{Sismanis}{2010}]%
        {Elo++}
\bibfield{author}{\bibinfo{person}{Yannis Sismanis}.}
  \bibinfo{year}{2010}\natexlab{}.
\newblock \showarticletitle{How {I} won the ``Chess Ratings -- {E}lo vs. the
  {R}est of the World'' Competition}.
\newblock \bibinfo{journal}{\emph{\url{arxiv.org/abs/1012.4571v1}}}
  (\bibinfo{year}{2010}).
\newblock


\bibitem[\protect\citeauthoryear{Stefani}{Stefani}{1977}]%
        {Stefani77}
\bibfield{author}{\bibinfo{person}{R. Stefani}.}
  \bibinfo{year}{1977}\natexlab{}.
\newblock \showarticletitle{Football and Basketball Predictions Using Least
  Squares}.
\newblock \bibinfo{journal}{\emph{IEEE Transactions on Systems, Man, and
  Cybernetics}}  \bibinfo{volume}{7} (\bibinfo{year}{1977}),
  \bibinfo{pages}{117--121}.
\newblock


\bibitem[\protect\citeauthoryear{Thurstone}{Thurstone}{1927}]%
        {Thurstone27}
\bibfield{author}{\bibinfo{person}{L.L. Thurstone}.}
  \bibinfo{year}{1927}\natexlab{}.
\newblock \showarticletitle{A law of comparative judgement}.
\newblock \bibinfo{journal}{\emph{Psychological Review}}  \bibinfo{volume}{34}
  (\bibinfo{year}{1927}), \bibinfo{pages}{278--286}.
\newblock


\bibitem[\protect\citeauthoryear{Tibshirani}{Tibshirani}{1996}]%
        {lasso}
\bibfield{author}{\bibinfo{person}{R. Tibshirani}.}
  \bibinfo{year}{1996}\natexlab{}.
\newblock \showarticletitle{Regression shrinkage and selection via the lasso}.
\newblock \bibinfo{journal}{\emph{Journal of the Royal Statistical Society,
  Series B}} \bibinfo{volume}{58}, \bibinfo{number}{1} (\bibinfo{year}{1996}),
  \bibinfo{pages}{267--288}.
\newblock


\bibitem[\protect\citeauthoryear{Xu, Huang, and Yao}{Xu et~al\mbox{.}}{2012}]%
        {MM12}
\bibfield{author}{\bibinfo{person}{Qianqian Xu}, \bibinfo{person}{Qingming
  Huang}, {and} \bibinfo{person}{Yuan Yao}.} \bibinfo{year}{2012}\natexlab{}.
\newblock \showarticletitle{Online crowdsourcing subjective image quality
  assessment}. In \bibinfo{booktitle}{\emph{ACM International Conference on
  Multimedia}}. \bibinfo{pages}{359--368}.
\newblock


\bibitem[\protect\citeauthoryear{Xu, Jiang, Yao, Huang, Yan, and Lin}{Xu
  et~al\mbox{.}}{2011}]%
        {mm11}
\bibfield{author}{\bibinfo{person}{Qianqian Xu}, \bibinfo{person}{Tingting
  Jiang}, \bibinfo{person}{Yuan Yao}, \bibinfo{person}{Qingming Huang},
  \bibinfo{person}{Bowei Yan}, {and} \bibinfo{person}{Weisi Lin}.}
  \bibinfo{year}{2011}\natexlab{}.
\newblock \showarticletitle{Random partial paired comparison for subjective
  video quality assessment via Hodgerank}. In \bibinfo{booktitle}{\emph{ACM
  International Conference on Multimedia}}. \bibinfo{pages}{393--402}.
\newblock


\bibitem[\protect\citeauthoryear{Xu, Xiong, Cao, and Yao}{Xu
  et~al\mbox{.}}{2016}]%
        {xu2016parsimonious}
\bibfield{author}{\bibinfo{person}{Qianqian Xu}, \bibinfo{person}{Jiechao
  Xiong}, \bibinfo{person}{Xiaochun Cao}, {and} \bibinfo{person}{Yuan Yao}.}
  \bibinfo{year}{2016}\natexlab{}.
\newblock \showarticletitle{Parsimonious Mixed-Effects HodgeRank for
  Crowdsourced Preference Aggregation}. In \bibinfo{booktitle}{\emph{{ACM}
  International Conference on Multimedia}}. \bibinfo{pages}{841--850}.
\newblock


\bibitem[\protect\citeauthoryear{Yu}{Yu}{2009}]%
        {Yu09}
\bibfield{author}{\bibinfo{person}{S. Yu}.} \bibinfo{year}{2009}\natexlab{}.
\newblock \showarticletitle{Angular Embedding: {F}rom Jarring Intensity
  Differences to Perceived Luminance}. In \bibinfo{booktitle}{\emph{IEEE
  Conference on Computer Vision and Pattern Recognition}}.
  \bibinfo{pages}{2302--2309}.
\newblock


\bibitem[\protect\citeauthoryear{Yu}{Yu}{2012}]%
        {Yu12}
\bibfield{author}{\bibinfo{person}{S. Yu}.} \bibinfo{year}{2012}\natexlab{}.
\newblock \showarticletitle{Angular Embedding: A Robust Quadratic Criterion}.
\newblock \bibinfo{journal}{\emph{IEEE Transactions on Pattern Analysis and
  Machine Intelligence}} \bibinfo{volume}{34}, \bibinfo{number}{1}
  (\bibinfo{year}{2012}), \bibinfo{pages}{158--173}.
\newblock


\bibitem[\protect\citeauthoryear{Zhang and Zhou}{Zhang and Zhou}{2014}]%
        {zhou2014}
\bibfield{author}{\bibinfo{person}{Min-Ling Zhang} {and}
  \bibinfo{person}{Zhi-Hua Zhou}.} \bibinfo{year}{2014}\natexlab{}.
\newblock \showarticletitle{A review on multi-label learning algorithms}.
\newblock \bibinfo{journal}{\emph{IEEE Transactions on Knowledge and Data
  Engineering}} \bibinfo{volume}{26}, \bibinfo{number}{8}
  (\bibinfo{year}{2014}), \bibinfo{pages}{1819--1837}.
\newblock


\end{thebibliography}

\end{document}